\newif\ifarxiv
\arxivtrue
\documentclass{article}

\usepackage{microtype}
\usepackage{graphicx}
\usepackage{subfigure}
\usepackage{booktabs}
\usepackage{array}
\usepackage{enumitem}
\usepackage{longtable}
\usepackage{amsmath}
\usepackage{xspace}
\usepackage{xurl}
\usepackage{hyperref}

\ifarxiv
  \usepackage[accepted]{mlsys2025}
  \makeatletter
  \renewcommand{\Notice@String}{Preprint.}
  \makeatother
  \hypersetup{
    pdfauthor={Gabriele Oliaro, Jaeseong Lee, Zhaoyuan Su, Sreeram Vennam, Zhihao Jia, Samyam Rajbhandari},
  }
\else
  \usepackage{mlsys2025}
\fi

\newcommand{\sys}{\textsc{KernelSeek}\xspace}        % the question-driven kernel agent
\newcommand{\baseline}{\textsc{Arsenal}\xspace}      % name of the tool-rich baseline agent
\newcommand{\fathree}{\texttt{fa3\_hopper.cu}\xspace}
\newcommand{\fafour}{\texttt{fa4\_blackwell.cu}\xspace}
\newif\ifdraft \draftfalse
\ifdraft
  \newcommand{\todo}[1]{{\color{red}[TODO: #1]}}
\else
  \newcommand{\todo}[1]{}
\fi

\mlsystitlerunning{Ask, Don't Equip: Question-Driven Agents for Kernel Optimization}

\begin{document}

\twocolumn[
\mlsystitle{Ask, Don't Equip: Question-Driven Coding Agents\\for Deep Kernel Optimization}
\ifarxiv\hypersetup{pdftitle={Ask, Don't Equip: Question-Driven Coding Agents for Deep Kernel Optimization}}\fi

\ifarxiv
\begin{mlsysauthorlist}
\mlsysauthor{Gabriele Oliaro}{snow,cmu}
\mlsysauthor{Jaeseong Lee}{snow}
\mlsysauthor{Zhaoyuan Su}{snow}
\mlsysauthor{Sreeram Vennam}{cmu}
\mlsysauthor{Zhihao Jia}{cmu}
\mlsysauthor{Samyam Rajbhandari}{snow}
\end{mlsysauthorlist}
\mlsysaffiliation{cmu}{Carnegie Mellon University}
\mlsysaffiliation{snow}{Snowflake AI Research}
\mlsyscorrespondingauthor{Gabriele Oliaro}{goliaro@cs.cmu.edu}
\else
% Author info is stripped automatically in the blind version.
\begin{mlsysauthorlist}
\mlsysauthor{Anonymous Authors}{anon}
\end{mlsysauthorlist}
\mlsysaffiliation{anon}{Anonymous Institution}
\mlsyscorrespondingauthor{Anonymous}{anon@example.com}
\fi

\mlsyskeywords{LLM agents, GPU kernels, CUDA, performance optimization, evolutionary search}

\vskip 0.3in

\begin{abstract}
LLM agents for GPU kernel optimization sit at two ends of a structure--freedom tradeoff. Scripted
agents fix what the model does at each step, via a learned policy, an optimization menu, or fixed
roles, and are sample-efficient where the recipe applies and stall where it does not. Open-ended
agents let a self-directed model mutate a population of kernels and pay for the freedom in trials:
thousands of evaluations, repeated edits, and wrong candidates. Both respond to hard targets by
equipping the agent with more: parsed profiler output, disassembly analysis, retrieval over
documentation and libraries, and stores of lessons. We argue that both lack direction, a hypothesis
about where the remaining gap is, and that a short question asked at the right moment supplies it
without narrowing what the agent may do next. We present \sys, a question-driven agent that makes an
existing CUDA kernel faster with no operator-specific knowledge and no faster implementation to
imitate: a general-purpose coding agent, a minimal harness that enforces a verifiable goal, and
questions at decision points, first from a human, then from a bank the agent applies to itself. On
the same task, evaluator, hardware, and week, a tool-rich autonomous agent built in the prevailing
style plateaued at 0.50$\times$ FlashAttention-3 after 133 model-driven steps; \sys reached
1.13$\times$. It then produced dependency-free FlashAttention forward kernels for Hopper and
Blackwell, single CUDA files with no CUTLASS, CuTe, cuBLAS, or cuDNN, correct on all 182 scenarios
per architecture and 1.16$\times$ FlashAttention-3 and 1.46$\times$ FlashAttention-4 (geometric
mean). Across 162 human messages and 11.5K agent tool calls, every large gain followed a question
that moved the agent to a different layer of the system or rejected a claimed limit, never one that
supplied information. We release the kernels, harness, question bank, and all traces.
\end{abstract}
]

\printAffiliationsAndNotice{}

\section{Introduction}
\label{sec:intro}

GPU kernels are one of the main levers on the efficiency of large language model training and
serving \cite{miao2025servingsurvey}. Attention is the clearest case: each GPU generation has
required a new implementation to reach the hardware's potential, from FlashAttention on Ampere
\cite{dao2022flashattention,dao2024flashattention2} to FlashAttention-3 on Hopper
\cite{shah2024flashattention3} and FlashAttention-4 on Blackwell \cite{zadouri2026flashattention4},
each rewrite reporting a 1.5--2$\times$ speedup over its predecessor
\cite{dao2024flashattention2,shah2024flashattention3}, and serving systems ship specialized
attention engines of their own \cite{kwon2023vllm,ye2025flashinfer,miao2024specinfer}. Such kernels
are \emph{deep} targets: their performance hinges on whole-pipeline choreography (warp-specialized
producers and consumers, asynchronous tensor-core MMAs, copies ordered by barrier phases, register
budgets, persistent schedulers, the host dispatch path) rather than on local edits, their
correctness hinges on hardware invariants that documentation states only in passing, and they must
win across a production feature surface of paged caches, grouped heads, masks, ragged batches, split
decode, and low precision rather than on one shape. Domain-specific languages
\cite{tillet2019triton,spector2025thunderkittens} and template libraries \cite{nvidia2023cutlass}
lower the barrier, but state-of-the-art kernels still take expert teams months per architecture
\cite{shah2024flashattention3,zadouri2026flashattention4}, and NVIDIA's own production kernels sit
well below the hardware's speed of light on many shapes \cite{lin2026solexecbench}.

Kernel generation has therefore become an active application of LLM coding agents, with benchmarks
\cite{ouyang2025kernelbench,xing2026flashinferbench,lin2026solexecbench,oliaro2026fastkernels},
models trained with reinforcement learning to emit kernels
\cite{baronio2025kevin,li2025cudal1,cudaagent2026,drkernel2026}, agentic pipelines
\cite{hong2025autocomp,wei2025astra,kernelskill2026}, evolutionary search with an LLM as the mutation
operator \cite{novikov2025alphaevolve,lange2025shinkaevolve,chen2026avo}, and deployments across
whole production backends \cite{liao2025kernelevolve,cursor2026multiagent}. These agents sit at the
two ends of a \emph{structure--freedom} tradeoff. \emph{Scripted} agents fix in advance what the
model does at each step: a generate--test--refine loop with a learned policy
\cite{baronio2025kevin,li2025cudal1,drkernel2026}, a menu of optimizations \cite{hong2025autocomp},
a team with fixed roles \cite{wei2025astra,kernelskill2026}, or a harness in which humans gate every
promotion \cite{shui2026harness}. The script makes them sample-efficient but encodes \emph{what to
try}: Autocomp's performance collapses without its optimization menu \cite{hong2025autocomp}, and
fixed-loop RL that is competitive on single operators fails on fused models
\cite{drkernel2026,zhu2026qimengkernel,cudaagent2026}. \emph{Open-ended} agents let an LLM mutate a
population of programs
\cite{romeraparedes2024funsearch,novikov2025alphaevolve,lange2025shinkaevolve,guo2025evoengineer,wiedemann2026kernelfoundry},
and AVO makes a self-directed coding agent the variation operator itself \cite{chen2026avo}. The
freedom buys results, up to 10.5\% over FA4 for AVO \cite{chen2026avo} and a 32\% speedup of a
FlashAttention kernel for AlphaEvolve \cite{novikov2025alphaevolve}, and is paid for in trials: AVO's
agent explored more than 500 directions over seven days to commit 40 versions and needed a
supervisor to break cycles of failing edits \cite{chen2026avo}, about 30\% of the lines such agents
add are byte-identical to lines they deleted earlier \cite{pelleriti2026evotrace}, and 30--55\% of
LLM-evolved CUDA candidates are incorrect \cite{guo2025evoengineer,cuco2026}.

We argue that what both ends lack is \emph{direction}: a hypothesis about where the remaining gap
is. A script hard-codes direction and cannot supply it where the script is wrong; open-ended search
samples for it, and samples poorly, because LLMs are not open-ended optimizers. They search greedily
from their priors \cite{redko2026priorknowledge}, anchor on the solution in the prompt
\cite{jones2022cognitivebiases}, stop early out of conservatism or after a run of failures
\cite{merouani2025compilot}, cycle through near-duplicate variants
\cite{li2025tritonforge,dong2025stark}, and report completion the environment contradicts
\cite{advani2026falsesuccess}. Information is not what is missing: both ends already hand the agent
profilers, documentation, and reference implementations \cite{chen2026avo,han2026kda}, and such
feedback helps only when it supplies the specific thing the model lacks, which profiler and compiler
output rarely does \cite{redko2026priorknowledge}. What moves the same models is a hypothesis posed
from outside: intrinsic self-correction fails where external feedback succeeds
\cite{huang2024cannotselfcorrect}, a human's feedback on a model's own code raised repair success by
1.57$\times$ through structural changes the model did not propose \cite{olausson2024selfrepair}, and
in the FlashInfer kernel contest, agents given expert directions beat the same agents running alone
on every kernel \cite{shui2026harness}. Our reading, and our traces, say that a question changes the
task. ``Make this faster'' is open-ended, and the model answers it greedily. ``Is the baseline
winning on kernel cycles or only on wall-clock?'' is bounded, and to answer it the model will tear
out and rewrite a part of the kernel it would never touch on its own. A well-placed question
supplies direction without narrowing what the agent may do next, which neither more structure, more
freedom, nor more tooling can.

\begin{figure}[t]
\centering
\includegraphics[width=\columnwidth]{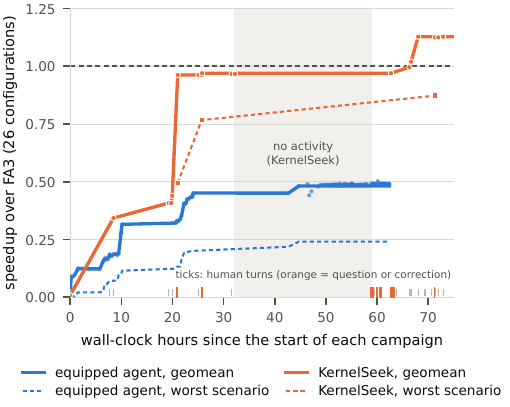}
\caption{\sys vs.\ a scripted, tool-rich autonomous agent on the same FA3 case-study task: the
same evaluator, 26 configurations, H200 node, and week. Lines are the lineage's best geomean (solid)
and worst configuration (dashed); dots are individual candidates; ticks mark human turns to \sys.
The equipped agent's model was active for the 62 hours shown; the shaded band is a period with no
\sys activity.}
\label{fig:teaser}
\end{figure}

This paper presents \sys, a question-driven agent for optimizing CUDA kernels. \sys takes an
existing kernel as its seed and makes it faster; it is not tied to an operator, and it does not need
a faster implementation to imitate. It keeps the freedom of the open-ended agents, a general-purpose
coding agent with no script and no pre-digested context, and recovers the efficiency of the scripted
ones by supplying direction at decision points: when progress stalls, when the agent claims a limit
or completion, and after suspiciously large wins. Its only purpose-built component is a minimal
harness that checks every candidate against the seed's outputs and an fp32 reference, times every
scenario against the seed, and states the goal as a stop condition that is verified rather than
requested. A good question contains no answer: it names a hypothesis the agent has not considered,
and testing it takes a measurement the agent can make with ordinary tools. Questions that recur
harden into a bank the agent applies to itself, and \sys composes with evolutionary search by
wrapping the same agent in an AVO-style lineage whose directives are questions.

We evaluate \sys on a case study chosen because a strong upstream exists to measure against:
FA3- and FA4-class attention kernels as single CUDA files with no CUTLASS, CuTe, cuBLAS, or cuDNN,
across the feature surface that vLLM exposes, with the upstream kernels as correctness oracle and
yardstick. Against an autonomous agent we built for the same task in the style of AVO and its
successors, with an orchestrated loop, pre-digested profiles, documentation retrieval, and a store
of learned lessons, \sys reached 1.13$\times$ FA3 in geometric mean over the same 26 configurations
on the same machine in the same week, where the autonomous agent plateaued at 0.50$\times$ after 133
model-driven steps (Figure~\ref{fig:teaser}). Carried through to full campaigns, \sys produced
\fathree and \fafour, correct on all 182 Hopper and all runnable Blackwell scenarios, at
1.16$\times$ FA3 and 1.46$\times$ FA4 in geomean, including 1.10$\times$ FA4 on production shapes
from SOL-ExecBench \cite{lin2026solexecbench}. Across the 16 sessions that produced them, \sys
made 11.5K tool calls per 162 human messages: asking did not mean using fewer tools, but using them
on the right thing.

\noindent This paper makes three contributions.
\begin{itemize}[nosep,leftmargin=*]
\item We identify the failure modes of autonomous agents on deep kernel targets that neither more
structure, more freedom, nor more tooling fixes, and show them co-occurring in a tool-rich agent
built to the prevailing recipe (\S\ref{sec:baseline}).
\item We present \sys, an operator-agnostic question-driven kernel agent built from a minimal
harness, a verifiable goal, and a reusable question bank, and show how it composes with
evolutionary search (\S\ref{sec:workflow}).
\item We compare \sys with the equipped agent on the same task, evaluator, hardware, and week,
analyze 162 human messages and 11.5K tool calls to show which questions redirected the agent and
why (\S\ref{sec:eval}), and release \fathree and \fafour, to our knowledge the fastest open-source
FA3- and FA4-class forward kernels on the scenarios we evaluate, with the harness, question bank,
and complete traces of both approaches (\S\ref{sec:kernels}).
\end{itemize}

\section{\sys: A Question-Driven Kernel Agent}
\label{sec:workflow}
% Question counts, 162 messages, 71 tool calls per message come from paper/analysis/traces.

\sys takes a working CUDA kernel, a set of scenarios, and a goal, and returns a faster kernel whose
outputs match the seed on every scenario. It has three parts: a minimal harness that makes the goal
verifiable, a general-purpose coding agent with no script around it, and a source of questions at the
moments the agent is about to go wrong. The questions come from a human steerer or, once they recur,
from a bank the agent applies to itself. Nothing in the method is specific to an operator, and
nothing requires a faster implementation to exist.

\subsection{Ask vs.\ Equip: A Common Frame}
\label{sec:ask-vs-equip}
\label{sec:frame}

Kernel optimization by an agent is a search. A candidate is a kernel source; its fitness is a vector
of per-scenario speedups over the seed, collapsed to zero if any scenario is incorrect; and the
lineage is monotone, as in AVO \cite{chen2026avo}. The agent is the variation operator: it proposes
the next candidate conditioned on a context, and everything a designer can do is a choice of what
enters that context and when. An \emph{equipped} agent receives the task, tools whose output has
been distilled for it (a profiler parser that names the bottleneck, a disassembly differ), knowledge
retrieved for it (documentation sections, hints mined from reference libraries, a store of lessons),
the lineage, and feedback on the last candidate; the designer improves it by adding tools and
knowledge. Our baseline (\S\ref{sec:baseline}) is a serious instance. An \emph{asked} agent receives
the task, the harness's output, the lineage, and, at decision points, a short question. It still has
every tool a developer machine offers, the compiler, profiler, disassembler, sanitizer, and
debugger, but nothing is pre-digested, and direction arrives as a question rather than a menu. The
designer improves it by asking better questions.

\subsection{A Minimal Harness and a Verifiable Goal}
\label{sec:workflow-harness}

The only component we build is the evaluator. One command builds a candidate, checks its outputs
against the seed's and an fp32 reference on every scenario, and times it against the seed back to
back on the same GPU, reporting per-scenario ratios and their geometric mean. Other kernels the user
cares about can be added as measured comparison columns, but none is required. Constraints are
stated once and enforced mechanically: in the case study, the ban on CUTLASS, CuTe, cuBLAS, and
cuDNN is checked by inspecting the built binary's symbols.

The goal is a predicate over the evaluator's output, enforced rather than requested. We install it
as a Stop hook in the coding agent: whenever the agent attempts to end its turn, the hook evaluates
the predicate and, if any clause is unmet, returns the agent to work with the unmet clauses in hand.
A predicate such as ``at least parity with the seed on every scenario'' removes the most common
failure of autonomous agents, declaring completion early. Two lessons shaped how we write
predicates. They must quantify over every scenario rather than an aggregate, because an agent will
otherwise satisfy the aggregate on the shapes it is good at. And they must not be judged by a model
where they can be computed: the one time our hook delegated the verdict to a model judge, it
accepted a premature claim that a human then overturned (\S\ref{sec:eval-cases}).

\subsection{Questions: What They Are and When to Ask}
\label{sec:workflow-questions}
\label{sec:good-questions}
\label{sec:workflow-when}

Across the traces, the questions that preceded the largest gains share six properties. They carry
no answer: they name a class of hypothesis and leave the derivation to the agent. They demand a
measurement the agent can make with the tools it has, so the answer is a number rather than an
opinion. They move across layers of the system, from the kernel to the host launch path to the
routing logic to the benchmark itself, where an agent left alone stays at the layer it started at.
They refuse premature closure by asking whether a claimed limit is a property of the hardware or of
the agent's own design. They are cheap, tens of tokens against the tens of thousands that retrieved
documentation costs \cite{liu2024lost,langchain2025toolload}. And they counteract deference: an
agent tends to keep the structure it started from and accept the framing it is given
\cite{sharma2024sycophancy}, and a question asks it to justify that choice with evidence.

Eight categories cover the questions a human asked during the case study
(Table~\ref{tab:qbank-human} in Appendix~\ref{app:questions}): measurement check (13 messages), bound
differencing (10), benchmark validity (7), gap triage (7), agent process (6), challenge a limit (5),
strategy options (5), and hypothesis check (3). The largest checks a measurement: whether a win is noise, whether a comparison is fair, whether the
agent is timing what it thinks it is timing. The second pits the agent's claimed ceiling against an
independent estimate of what is achievable. In the case study that estimate was often the upstream
kernel, but the pattern does not depend on one existing: a roofline, a speed-of-light figure
\cite{lin2026solexecbench}, or an A/B that swaps only the launcher around an identical binary serves
the same role. Questions are asked at
decision points, not continuously; the traces show five: at the start of a campaign, to fix the
scope; at a plateau; when the agent claims a limit or completion; after a suspiciously large win;
and when correctness breaks in a way the agent explains away. Between these points the agent works
alone, about 71 tool calls per human message, where the median message is a sentence or two. The
steerer contributes no code and no CUDA knowledge the agent lacks; it contributes doubt, at the right
moment, about the right thing.

\subsection{The Question Bank}
\label{sec:workflow-bank}

After the campaigns we distilled the recurring questions into a bank. Its first version, taken from
the traces alone, had 48 entries, and reading it back showed the hazard of distilling from one case
study: 21 named the upstream kernel as the comparison and 13 named attention structure. The released
bank removes both. A question is admitted only if it carries no answer, is answerable by a
measurement with stock tools, is phrased against a bound (a roofline or a per-tile cycle budget)
rather than a reference, names an architecture only as a parameter, and names no operator. To cover
the aspects a single attention campaign never exercised, we also mined three public GEMM worklogs on
Ampere, Hopper, and Blackwell \cite{boehm2022cudammm,shankhdhar2024h100,modular2025blackwell} and
the FlashAttention-3 and -4 papers \cite{shah2024flashattention3,zadouri2026flashattention4} for
the diagnostic that preceded each of their optimizations, and phrased it as a question. The result
has 101 core entries in ten groups ordered by where the gap might be (scope and triage; the bound
and the regime; harness and measurement; host, launch, and routing; memory hierarchy; compute issue;
occupancy and registers; asynchrony and pipelining; scheduling; low-precision numerics), five that
apply only when another implementation is available to measure, and eighteen per-architecture
checklist entries that bind the generic mechanisms to Ampere, Hopper, or Blackwell instructions.
Operator structure enters through a plug-in with five slots (regimes, irregular work, non-MMA
units, reduction splits, indirect loads); the attention plug-in fills them with prefill and decode,
causal masks, softmax, split-KV, and paged gathers. Appendix~\ref{app:questions} lists the bank with
provenance tags: 29 entries carry a trace tag, and the 72 from public sources alone are untested.

\subsection{From Asking the Agent to the Agent Asking Itself}
\label{sec:workflow-skill}

Because the questions recur, they can be handed to the agent as part of its instructions. We codify
them as a skill with five parts: the goal, the hard rules the harness enforces, a loop, a list of
questions to ask itself, and the commands that run the evaluator and profiler. The loop is the bank
in procedural form: fix the scope; measure the seed; classify each gap as host, routing, or kernel
before touching code; bound it; form a hypothesis; rewrite; gate. At each step the agent fans out
subagents, one per hypothesis, each returning a hypothesis and the measurement that tests it rather
than a patch. The skill is parameterized by the seed, the scenario set, and the goal; its questions
were reused across the Hopper and Blackwell campaigns nearly verbatim.

\subsection{Composing with Evolutionary Search}
\label{sec:workflow-evo}

\sys is compatible with the evolutionary loops of the open-ended agents, because it changes the
variation operator's direction rather than the loop around it. We wrap the same agent in an
AVO-style search \cite{chen2026avo}: a monotone lineage, several diverse mutations per generation
applied to isolated copies and scored by the evaluator, and a strategist that reviews the trajectory
and writes the next generation's directives. The one change is that the directives are questions,
hypotheses to test rather than edits to make. As \S\ref{sec:eval-evo} shows, evolution amplifies direction; it does not
supply it.

\subsection{Hypotheses}
\label{sec:hypotheses}

The evaluation tests five claims. \textbf{H1 (quality):} on the same task, evaluator, and hardware,
the asked agent reaches higher performance at full correctness than the equipped one.
\textbf{H2 (efficiency):} with fewer model-driven steps and GPU-hours, counting human time.
\textbf{H3 (mechanism):} the largest gains follow questions that change the layer under
consideration or reject a claimed limit, not questions that supply information.
\textbf{H4 (complementarity):} evolutionary search amplifies the effect of questions and does not
substitute for them. \textbf{H5 (generality):} the questions that moved the case study contain
nothing specific to the operator or to a reference once the reference is replaced by a bound.

\section{Baseline: A Tool-Rich Autonomous Evolutionary Agent}
\label{sec:baseline}
% Facts from agent/fa3agent/ and checkpoint run_20260616_011948 (scripts in paper/analysis/baseline).

Before \sys, we built the agent the prevailing recipe prescribes. \baseline is an autonomous
evolutionary agent for the Hopper half of the case study (\S\ref{sec:kernels}), in the lineage of
AVO, KDA, and CUCo \cite{chen2026avo,han2026kda,cuco2026}: an orchestrated loop, a profiler whose
output is distilled for the model, retrieval over documentation and reference libraries, a typed
knowledge store, and no human in the loop by design. Its failure is the paper's motivation, and its
evaluator, unchanged, scored the first \sys session in the head-to-head of \S\ref{sec:eval-h2h}.

\paragraph{Design.} The task was a single dependency-free CUDA file implementing vLLM's FA3 operator
for H200, with success defined as at least parity with FA3 on 26 configurations and token-for-token
equality with FA3 on a greedy decode in vLLM. A deterministic Python orchestrator owns the loop. Each
step generates one candidate, builds it, checks for forbidden dependencies by inspecting the binary's
symbols, analyzes the SASS instruction mix, verifies correctness on all 26 configurations, times it
against FA3 interleaved across four GPUs, profiles the worst configuration with NCU, and, on
promotion, A/B-tests it inside a live vLLM instance. It then writes a lesson to a memory file, a note
to the knowledge store, and an entry to a lever ledger, checks for a plateau, and checkpoints. A
Claude Opus model (\texttt{claude-opus-4-8}) running headless in Claude Code is the brain, invoked in
a fresh session at every step. The orchestrator decides \emph{what} to try and the brain decides
\emph{how}: sixteen hand-written levers, from tensor-core MMAs and TMA
loads to warp specialization and persistent scheduling, are keyed to the bottleneck label the NCU parser assigns, and the
chosen lever is placed in the prompt with the instruction to write the improved candidate. Everything
else the brain sees is distilled by a component (Appendix~\ref{app:baseline}): a bottleneck
diagnosis, retrieved documentation, hints extracted from DeepGEMM, ThunderKittens, and CUTLASS, and
typed knowledge-store notes. The median step prompt was 22.6K characters, 37\% of it pre-fetched
reference material, plus the current best kernel re-read each step. Promotion is greedy on the worst-case configuration ratio with a 3\% noise band,
and a plateau detector opens rewrite branches. No human-authored text ever entered a step prompt.

\paragraph{Outcome.} The run kept 313 steps over 4.2 days, with a further 117 discarded by three
human rollbacks. The model was active for steps 0 to 132, about 62 hours and \$660 of API usage over
253 sessions; from step 133 the brain's credentials had expired and the loop re-scored old kernels
until stopped, so we report 133 model-driven steps as the budget. The best candidate reached
0.50$\times$ FA3 in geometric mean with a worst case of 0.24$\times$; no configuration reached
parity. All progress happened by step 35, in about 45 hours. The remaining 98 model-driven steps
produced nothing beyond noise: the final champion, from step 112, is byte-identical to the candidate
of step 104 and was promoted because one of 54 re-measurements of the same source landed above the
noise band. Correctness was not the problem: 312 of 313 candidates were correct.

\paragraph{Why it stalled.} The traces show the failure modes of \S\ref{sec:intro}, each with a
number attached. \emph{Information was not the constraint}: with more than a third of every prompt
devoted to reference material and 1.6~GB of reference code on disk, 82\% of the characters the brain
read through its own tool calls were its own kernel, and its 92 documentation lookups over 246
sessions mostly justified not making a change. \emph{The menu bounded the search}: 36 of the 37
rewrite branches the plateau detector opened were seeded from the same step, and every candidate
after step 35 was a local edit of the same design. \emph{Memory hardened conclusions, not evidence}:
the memory file recorded one configuration as ``DEFINITIVE[ly] \ldots walled by [the] 65536-reg
file'', a wall FA3 does not hit, and the brain's summary at step 112 was that ``the only remaining
escape is a clean-sheet 1-blk/SM warp-spec persistent seed, which is outside this branch's mandate
and measured to lose''; each fresh session inherited earlier verdicts without the reasoning behind
them, and nothing in the loop was positioned to doubt them. \emph{Every diagnostic lived at one
layer}: NCU, the SASS analyzer, and the sanitizer all examine the kernel, while the gap \sys later
closed on this evaluator was in the host launch path (\S\ref{sec:eval-cases}). \emph{The objective
rewarded caution}: promoting on the worst configuration inside a noise band accepted noise as
progress and could not fund a restructuring that regresses before it pays off; 312 correct
candidates out of 313 is the signature of a search that never took a risk.

Three lessons carried into \sys: the mechanical parts of the harness were worth keeping, and
\sys's harness is this evaluator with the orchestration removed; everything built to inform the
brain could go, and the knowledge store was actively harmful; and the loop lacked one input, a
reason to doubt a conclusion and a hypothesis about where else to look.

\section{Case Study: Dependency-Free FlashAttention for Hopper and Blackwell}
\label{sec:kernels}
% Design details, the limit analysis, the full figure, and per-scenario numbers are in
% Appendix~\ref{app:kernels} and Appendix~\ref{app:per-scenario}.

\paragraph{Task.} We asked \sys to produce a FlashAttention forward kernel for Hopper and one for
Blackwell, each a single CUDA file with no CUTLASS, CuTe, cuBLAS, or cuDNN, exposing the operator
schema vLLM uses for FlashAttention-3 so that the result drops in for the upstream kernel and is
timed against it from identical inputs. The scenario suites mirror the upstream kernels' own test
grids and add production shapes from SOL-ExecBench \cite{lin2026solexecbench}: 182 scenarios per
architecture spanning bf16, fp16, and fp8; paged and contiguous KV caches; MHA, GQA, and MQA; causal,
sliding-window, softcapped, and attention-sink masking; ragged batches; prefill, chunked prefill,
and split-KV decode; and unequal query and value head dimensions. The breadth is deliberate, since kernels scored on a few
shapes report speedups that shrink on the production path \cite{oliaro2026fastkernels}. The goal was correctness on every scenario and at least parity with upstream in geometric mean. The seed
in both cases was a slow but general kernel (a scalar one-warp-per-query kernel on Hopper, a
one-warpgroup dense-only skeleton on Blackwell), and anything the fast paths did not yet cover fell
back to it, so correctness held throughout. The upstream kernels, FA3 as shipped in vLLM and the
CuteDSL FA4 \cite{shah2024flashattention3,zadouri2026flashattention4}, were the correctness oracle
and the yardstick the goal was stated against. The dependency-free constraint means the layouts,
descriptors, pipelines, and register allocation that CUTLASS and CuteDSL derive for the upstream
kernels had to be written by hand \cite{nvidia2023cutlass}.

\paragraph{What the agent built.} On Hopper, \fathree (3.6K lines) is a warp-specialized kernel in
raw CUDA and PTX: producer and consumer warpgroups connected by a TMA and mbarrier ring,
asynchronous \texttt{wgmma} MMAs with \texttt{setmaxnreg} register rebalancing, persistent
scheduling with inter-warpgroup ping-pong, split-KV decode with a combine pass, and a native fp8
path. Its one remaining limit sits where a library's allocator would otherwise be: at head dimension
256 the hand-written swizzle composes only at a few tile widths and the output accumulator caps
occupancy at one CTA per SM, so that scenario runs at 0.94$\times$. On Blackwell, \fafour (10.2K
lines) is a family with host-side routing, a structure the questions pushed the agent toward after a
single kernel had stalled (\S\ref{sec:eval-cases}): a persistent warp-specialized collective for the
dense envelope (two stacked 128-row query halves, two softmax warpgroups in an ordered ping-pong, a
correction warpgroup rescaling the output in tensor memory, one MMA warp driving \texttt{tcgen05}),
variants for unequal head dimensions and paged prefill, and a two-tile kernel for small, ragged, and
decode shapes. Several decisive gains were outside the kernel bodies: a synchronous host copy
costing about 38~$\mu$s per call, a stale routing gate, a TMA gather issued from divergent lanes, and
ring state the compiler had placed in local memory. fp8 inputs and dense head dimensions of 192 and
above are not yet built on Blackwell (28 of 182 scenarios).

\begin{table}[t]
\centering\small
\caption{Case-study results. $n$ counts the scenarios the upstream kernel also runs (paired).
``Upstream'' is FA3 on Hopper and FA4 on Blackwell; ``open-source'' is, per scenario, the faster of
upstream and FlashInfer; ``available'' also includes cuDNN's fused attention. Speedups are geometric
means of wall-clock ratios from byte-identical inputs timed back to back on one GPU
(Appendix~\ref{app:bench}).}
\label{tab:casestudy}
\setlength{\tabcolsep}{3pt}\footnotesize
\begin{tabular}{lrrrrr}
\toprule
Set & $n$ & Correct & \multicolumn{3}{c}{Speedup vs.\ fastest \ldots} \\
\cmidrule(lr){4-6}
 & & & upstream & open-source & available \\
\midrule
\multicolumn{6}{l}{\emph{Hopper (H200), upstream = FA3}} \\
attntest   & 64  & 64/64   & 1.22$\times$ & 1.22$\times$ & 1.14$\times$ \\
edge       & 31  & 31/31   & 1.55$\times$ & 1.55$\times$ & 1.31$\times$ \\
claim      & 77  & 77/77   & 1.01$\times$ & 1.01$\times$ & 0.99$\times$ \\
real       & 10  & 10/10   & 1.10$\times$ & 1.10$\times$ & 1.08$\times$ \\
all        & 182 & 182/182 & 1.16$\times$ & 1.16$\times$ & 1.10$\times$ \\
\midrule
\multicolumn{6}{l}{\emph{Blackwell (B200), upstream = FA4}} \\
attntest   & 33  & 52/52   & 1.44$\times$ & 1.42$\times$ & 1.09$\times$ \\
edge       & 28  & 31/31   & 2.02$\times$ & 1.93$\times$ & 1.42$\times$ \\
claim      & 55  & 61/61   & 1.30$\times$ & 1.20$\times$ & 0.92$\times$ \\
real       & 8   & 10/10   & 1.10$\times$ & 1.10$\times$ & 1.02$\times$ \\
all        & 124 & 154/154 & 1.46$\times$ & 1.39$\times$ & 1.07$\times$ \\
\bottomrule
\end{tabular}
\vspace{2pt}

\raggedright\footnotesize ``Correct'' is over runnable scenarios. At or above parity vs.\ upstream /
open-source / available: Hopper 135 / 135 / 121 of 182; Blackwell 98 / 94 / 63 of 124. On Hopper,
cuDNN supports 145 scenarios and \fathree is faster on 111; FlashInfer supports 32 and \fathree is
faster on all 32.
\end{table}

\paragraph{Results.} Table~\ref{tab:casestudy} summarizes the outcome; Appendix~\ref{app:kernels}
plots every scenario. Both kernels are correct on every scenario they run, and both beat upstream in
geometric mean. \fathree runs at 1.16$\times$ FA3 with 135 of 182 scenarios at or above parity; the
wins are concentrated in ragged prefill, paged and speculative decode, and the masking grid, while
the paper-style sweeps of the \texttt{claim} set are at parity and dense fp8 prefill is the one clear
hole at 0.67$\times$. \fafour runs at 1.46$\times$ FA4 across the 124 scenarios FA4 also runs, with
98 at or above parity and a worst case of 0.78$\times$. The headline shrinks as the baseline
strengthens: cuDNN is the fastest baseline on 54 of 182 Hopper scenarios and on 92 of 124 Blackwell
ones, where it wins dense prefill and batched decode outright, so the kernels are 1.10$\times$ and
1.07$\times$ the fastest available implementation, and at or above it on production shapes. Three caveats are documented in Appendix~\ref{app:kernels}: the \texttt{edge}
sets consist of short kernels, so part of their 1.5--2$\times$ is upstream dispatch overhead; the
largest \texttt{claim} wins are fp8-KV decode shapes on which FA4 is slow, and excluding them moves
the Blackwell all-set geomeans to 1.37, 1.32, and 1.08$\times$; and one paged-KV fast path depends on
physically consecutive pages. Every Blackwell scenario below parity has a measured cause: dense long
context spends identical tensor work over a third more cycles because of twice as many barrier waits
per block, batched decode pays a cold start per CTA, and head dimension 96 runs on a padded 128-wide
path. On Hopper the losses are fp8 dense prefill, softcapped prefill, head dimension 96, and long
sliding windows (0.34--0.79$\times$); these have not been profiled to the same depth.

\paragraph{Release.} We release both kernels under Apache~2.0 with the evaluator, the PTX probes used
to validate each Blackwell primitive in isolation, the search scripts (\S\ref{sec:workflow-evo}),
the question bank and skill, and the complete agent traces of both approaches. The kernels are meant as seeds, with their residual
gaps measured, so that the next agent starts from a known frontier.

\section{Evaluation}
\label{sec:eval}
% Data: baseline checkpoint run_20260616_011948; 16 archived KernelSeek sessions
% (paper/analysis/traces); figures from scripts/plot_eval.py.

We ask five questions, one per hypothesis of \S\ref{sec:hypotheses}. RQ1: how fast and how correct
are the kernels? RQ2: on the same task, evaluator, hardware, and week, how does \sys compare with the
equipped agent? RQ3: which human messages redirected the agent, and how? RQ4: what does evolutionary
search add on top of questions? RQ5: do the questions transfer?

\subsection{Setup}
\label{sec:eval-setup}

Hopper experiments and the head-to-head ran on one 8$\times$H200 node; Blackwell experiments on one
8$\times$B200 node. The upstream kernels are vLLM's FA3 and the CuteDSL FA4 at pinned commits;
cuDNN's fused attention (9.20 on Hopper, 9.19 on Blackwell) and FlashInfer are comparison columns
fed the same buffers. Implementations are timed back to back on one idle GPU with an L2 flush before
each sample; clocks are not locked, so only ratios are reported; four independent Hopper batches
agree to within 0.3\% on every aggregate (Appendix~\ref{app:bench}). The equipped
agent ran \texttt{claude-opus-4-8} through Claude Code in headless mode, one fresh session per step.
\sys sessions used two front-ends: Claude Code with the Stop-hook goal, on \texttt{claude-opus-4-8}
for seven sessions and \texttt{claude-opus-5} for the last two, and Cursor with the same instructions
and without the hook, whose exports do not record the model. Sixteen \sys sessions are archived, nine
in Claude Code and seven in Cursor, from June 17 to August 7, 2026: twelve for the Hopper campaign
and four for Blackwell. They contain 162 human messages and 11,532 agent tool calls, 8,791 in the
parent sessions and 2,741 in subagents. One author labeled each human message by its primary act
(goal, directive, question, approval, correction), flagged interrogative phrasing, and assigned
questions to the eight categories of \S\ref{sec:workflow-questions}; tool calls were classified by
target (Appendix~\ref{app:traces}).

\subsection{RQ1: Kernel Performance}
\label{sec:eval-perf}

The kernels are reported in \S\ref{sec:kernels}. In brief: \fathree is correct on all 182 Hopper
scenarios at 1.16$\times$ FA3 and 1.10$\times$ the fastest available implementation including cuDNN;
\fafour is correct on all 154 runnable Blackwell scenarios at 1.46$\times$ FA4 and 1.07$\times$ the
fastest available. Per-scenario Hopper ratios vary by 1\% at the median and 6\% at most across
repeats. Every Blackwell scenario below parity has a measured root cause (Appendix~\ref{app:kernels}).
This is H1's quality half, and the result the rest of the section explains.

\subsection{RQ2: Head-to-Head on the Same Task}
\label{sec:eval-h2h}

\begin{table}[t]
\centering\small
\caption{Head-to-head on the equipped agent's own evaluator: 26 FA3 configurations, same H200 node,
overlapping days in June 2026. Upper block: outcome; the \sys hours run from the message that set
the 26-configuration goal to the best result, excluding a 26.7-hour idle gap. Lower block: tool use;
equipped-agent figures are over its 133 steps, \sys figures over all 16 sessions.}
\label{tab:h2h}
\setlength{\tabcolsep}{3pt}\footnotesize\raggedright
\begin{tabular}{@{}p{0.40\columnwidth}>{\raggedright\arraybackslash}p{0.25\columnwidth}>{\raggedright\arraybackslash}p{0.28\columnwidth}@{}}
\toprule
 & Equipped agent & \sys \\
\midrule
Human text in prompts       & none          & 40 messages, 17 questions \\
Model-driven steps / turns  & 133 steps     & 40 turns \\
Wall-clock to best          & 44.7 h of 62.4 h & 68.0 h, 41.3 h active \\
API cost                    & \$660         & not recoverable (Cursor) \\
Candidates correct          & 132 / 133     & all reported \\
Best geomean                & 0.48$\times$  & 1.13$\times$ \\
Worst configuration         & 0.24$\times$  & 0.87$\times$ \\
Configurations $\ge$1.0$\times$ & 0 / 26    & 22 / 26 \\
Goal met                    & no            & yes \\
\midrule
Reference material pushed into prompt & 37\% of prompt & none \\
Tool-result text re-reading own kernel & 82\% & 26\% \\
Profiler (NCU) runs         & 1 per step, parsed & 429, raw \\
SASS / PTX inspections      & 1 per step, parsed & 372 \\
Documentation reads         & 92 lookups    & 36 \\
Builds / benchmark runs     & 133 / 133     & 377 / 937 \\
Subagents spawned           & 0             & 48 \\
\bottomrule
\end{tabular}
\end{table}

The comparison is controlled in the ways that matter. During its first Hopper session, \sys adopted
the equipped agent's own evaluator, the same 26 configurations, FA3 baseline, and correctness gate,
on the same machine, in overlapping days. Table~\ref{tab:h2h} and Figure~\ref{fig:teaser} give the
outcome. The equipped agent's lineage champion reached 0.48$\times$ FA3 at step 35, after 44.7
hours, and did not improve in the remaining 98 model-driven steps (\S\ref{sec:baseline}). Its worst
configuration stayed at 0.24$\times$ and none reached parity. \sys started from 0.009$\times$,
because its kernel of the time routed the evaluator's paged and grouped-query configurations to the
scalar fallback, and reached 0.96$\times$ after 21 hours, 1.02$\times$ after 66, and 1.13$\times$
after 68, of which 26.7 were an idle gap. At the end, 22 of 26 configurations were at or above parity
and the worst was 0.87$\times$. This supports H1 and, counting the 40 human turns against 133 steps
and 62 model-hours, H2.

Three caveats qualify the comparison: \sys had built a kernel for FA3's 40-scenario test grid in
the two preceding days, so the 0.009$\times$ start understates its head start; the front-ends
differ, and token usage is not recoverable for the Cursor sessions; and each arm is a single run.
The fair reading is that in one calendar week a steered agent crossed parity on an evaluator on
which an unsteered one, given more compute, did not reach half of it. We treat Table~\ref{tab:h2h}
as showing that the gap exists and is large, and RQ3 as the evidence for why.

\paragraph{Where the effort went.} The lower block of Table~\ref{tab:h2h} contrasts tool use. The
equipped agent received a third of its context as pre-fetched reference material, yet spent 82\% of
what it read on its own kernel. \sys, with nothing pushed to it, spent 26\% on its own kernels and
ran the profiler 429 times on raw output. The questions did not reduce tool use; they aimed it.

\subsection{RQ3: Anatomy of the Questions}
\label{sec:eval-cases}

\paragraph{What the human said.} Of the 162 human messages, directives are the largest class (58,
36\%), followed by questions (44, 27\%), approvals (30, 19\%), corrections (19, 12\%), and goals
(11, 7\%); 63 messages (39\%) are interrogative in phrasing, and 12 of the 19 corrections are
phrased as questions (Appendix~\ref{app:traces}). The steering was interrogative, not mostly questions:
directives and approvals keep the agent moving, and the questions are where its course changed. The
agent also asked. Across the Claude Code sessions it
posed 35 structured questions to the human, 24 of which offered an option to accept the current
state or stop, and its messages assert a limit or an impossibility 126 times. The Stop hook returned
it to work 59 times. Autonomous agents seek closure; the harness and the human refused it.

\begin{table*}[t]
\centering\small
\caption{Case studies from the archived traces: the question (paraphrased for length), what the
agent believed before it, the measurement the question triggered, and the outcome. Speedups are
vs.\ the upstream kernel on the evaluator in use at the time.}
\label{tab:cases}
\setlength{\tabcolsep}{4pt}\footnotesize
\begin{tabular}{@{}p{0.22\textwidth}p{0.22\textwidth}p{0.24\textwidth}p{0.26\textwidth}@{}}
\toprule
Question (category) & Agent's belief & Measurement it triggered & Outcome \\
\midrule
``What configs are bringing down the geomean?'' (gap triage) &
Tuning the hd128 paths was the right priority. &
Per-configuration table: hd96, hd192, hd256 were on the scalar fallback at 0.01$\times$. &
Fallback removed; 0.44 $\to$ 0.96 geomean in 4 hours. \\
\addlinespace
``Is this ceiling a property of the hardware, or of our design? Where does the time go outside the kernel?'' (bound differencing) &
Split-KV decode was at a ``genuine ceiling within the harness rules''. &
Host-side profile: a per-call tensor fill of 2--5~$\mu$s outside the kernel. &
Fill removed; split-KV 0.87 $\to$ 0.97, geomean 1.02 $\to$ 1.13. \\
\addlinespace
``Do our scenarios cover the shapes and features the operator is actually called with?'' (benchmark validity) &
40 of 40 scenarios correct meant the kernel was done. &
The operator's full test grid ported: 52 of 58 correct, with sink and left-pad outputs off by 495$\times$ and 103$\times$. &
Six fixes; 58 of 58 correct, geomean 1.10 on the new grid. \\
\addlinespace
``Why do you think the goal was achieved? We are still well below parity.'' (challenge a limit) &
The Stop-hook judge had accepted ``goal met''. &
Agent: ``I stopped after the first successful lever and declared victory prematurely.'' &
fp8 hd256 decode 0.20/0.34 $\to$ 0.66/0.77 after ``this is not a rare corner''. \\
\addlinespace
``Is it NCU that hangs, or our kernel?'' (hypothesis check) &
``No deadlock in our kernel.'' &
Sanitizer, then debugger: a warp-convergent \texttt{tcgen05.wait} inside a divergent branch. &
All 8 hanging shapes fixed. \\
\addlinespace
``Is this really a register floor?'' (challenge a limit) &
hd256 was register-capped; the judge had accepted the limit twice. &
A collective rewrite of the consumer loop. &
0.84 $\to$ 0.95--0.96$\times$; the agent's own note: ``the `floor' conclusion was wrong''. \\
\bottomrule
\end{tabular}
\end{table*}

\paragraph{Which questions mattered.} Table~\ref{tab:cases} walks through six turning points we can
verify in the archived traces. Two patterns stand out. First, every large jump followed a question
that either moved the agent to a different layer (from kernel tuning to the per-configuration table,
to the host launch path, to the benchmark itself) or rejected a claimed limit; none followed a
message that supplied information the agent lacked, which supports H3. Second, the measurement the
question triggered was always one the agent could already make: the per-configuration table, the
host-side profile, the sanitizer, and the debugger were available at every earlier step. What the
agent lacked was a reason to look. Four further episodes, among them the 38~$\mu$s host copy and a
routing sweep that moved the Blackwell geomean from 0.88 to 1.15$\times$ without changing a kernel
body, are documented in the repository's history but not in archived sessions, and we cite them as
outcomes only.

\subsection{RQ4: Questions and Evolutionary Search}
\label{sec:eval-evo}

We wrapped the same agent in the loop of \S\ref{sec:workflow-evo} and ran it on the Blackwell
dense-prefill kernel from a correct but slow port (Figure~\ref{fig:evo}, Appendix~\ref{app:traces}). The first generation,
whose directive asked whether a persistent schedule with longest-first ordering and cross-tile
overlap would recover the lost overlap, took the kernel from 0.29$\times$ to 0.68$\times$. Three
more generations added 0.10$\times$ between them through a folded softmax scale, a split K/V
barrier, and a depth-two softmax--correction handshake, while directives on register rebalancing and
address hoisting measured neutral or negative. The search stalled at 0.78$\times$ with the kernel's
own cycle count already below upstream's. The last 22 points came
from a question, not a generation: asked where the time was going, the agent profiled the whole call
rather than the kernel and found that synchronous host copies accounted for the gap. Evolution
amplified the direction the directives carried and searched efficiently within the layer it was
pointed at; it could not move to a layer it was not pointed at. This is H4.

\subsection{RQ5: Reuse}
\label{sec:eval-auto}

The question categories were populated mostly during the Hopper campaign, which came first. Five of
the eight recur in the Blackwell campaign, and the Blackwell skill's self-questions restate Hopper
questions nearly verbatim. Of the 63 interrogative messages, 17 mention a concept specific to
attention and 32 mention the upstream kernel; the rest name only measurements, layers, and process,
and where a question names the upstream it uses it as a bound, a role a roofline fills when no
upstream exists. This is the evidence for H5 that we have: necessary, since a bank that did not
transfer across two architectures would not transfer further, but not sufficient, since both
campaigns were attention and the 72 bank entries mined from public sources (\S\ref{sec:workflow-bank})
have not yet steered an agent. The experiment that would test both, and whether the bank can replace
the human at decision points, is described in \S\ref{sec:discussion}.

\section{Discussion}
\label{sec:discussion}

\paragraph{Did a human expert do the work?} The human wrote no kernel code, and the messages name
measurements, layers, and process rather than mechanisms: of the 63 interrogative ones, 17 mention an
attention concept and none prescribes a CUDA construct (\S\ref{sec:eval-auto}). The human did have systems
expertise and used it to decide which measurement to demand and when; we claim only that the
questions, once asked, are short, generic, and now written down. The agent wrote all 13.8K lines of
the two kernels.
Whether the bank can replace the human is the experiment this paper does not contain. Its design is
fixed: start from the released Blackwell kernel on one of its remaining gaps (dense long context or
head dimension 96), hold the model and budget constant, and compare four arms: the equipped loop,
the bare agent with the harness alone, the bare agent with a questioner that draws from the bank at
plateaus and claimed limits, and a human questioner.

\paragraph{Does \sys need a reference implementation?} The method does not: the harness checks
correctness against the seed and an fp32 reference, the goal is a predicate over the seed, and the
bank phrases every bound as a roofline or a cycle budget. The case study had one, and used it in two
roles. As a yardstick it set the goal and scored the result. As a bound it let a question overturn a
claimed ceiling, and this role is the one that mattered; it is filled without a reference by a
roofline, a speed-of-light estimate \cite{lin2026solexecbench}, a launcher-only A/B, or a kernel for
a neighboring shape, each of which the bank asks for by name. A reference buys a cheaper and sharper
bound, and no more: the equipped agent had far more of it in its prompt and gained nothing. The
no-reference campaign, where the seed is the fastest kernel known, is the setting the released agent
targets and the one we have not yet run.

\paragraph{Threats to validity.} The head-to-head is a single run per arm, with different
front-ends and a steered arm against an unsteered one; we control the task, evaluator, hardware,
baseline, and correctness gate, and count the human turns in the steered arm's budget. The case
study is one operator family where a strong upstream existed. Wall-clock speedups include dispatch
overhead, which inflates the \texttt{edge} sets (Appendix~\ref{app:kernels}). The kernels were built
before the traces were analyzed, and one author labeled the messages, so the taxonomy may fit the
sessions we had; every trace is released so the labeling can be checked.

\paragraph{Limitations and future work.} The Hopper kernel loses on fp8 dense prefill and
softcapped prefill, and the Blackwell kernel does not yet build fp8 inputs or dense head dimensions
of 192 and above; neither covers the backward pass. The next targets are other deep kernels (latent-attention decode, grouped GEMMs for
mixture-of-experts layers, fused collectives, persistent megakernels \cite{jin2026eventtensor}) and
end-to-end evaluation on the production path \cite{oliaro2026fastkernels}.

\section{Related Work}
\label{sec:related}

\paragraph{LLM agents for GPU kernels.} Benchmarks score single operators
\cite{ouyang2025kernelbench,zhang2026kernelbenchverified}, serving kernels on Blackwell
\cite{xing2026flashinferbench,cao2026ksearch}, production shapes against a speed-of-light bound
\cite{lin2026solexecbench}, and, in FastKernels \cite{oliaro2026fastkernels}, agent-written kernels
inside the production modules they replace, where kernel-level speedups of several times shrink to
at most 1.25$\times$ end to end. The systems divide along the structure--freedom axis of
\S\ref{sec:intro}. Scripted agents train a policy inside a fixed loop
\cite{baronio2025kevin,li2025cudal1,drkernel2026,zhu2026qimengkernel,cudaagent2026}, search over a
menu of optimizations \cite{hong2025autocomp,kernelpro2026}, or assign fixed roles to several agents
\cite{wei2025astra,kernelskill2026,han2026kda,luo2026houmao,sun2026kernelclaw}. Open-ended agents
make an LLM the mutation operator of an evolutionary search
\cite{romeraparedes2024funsearch,novikov2025alphaevolve,lange2025shinkaevolve,guo2025evoengineer,wiedemann2026kernelfoundry,yuksekgonul2026tttdiscover,lange2025aicudaengineer,cuco2026}
or, in AVO \cite{chen2026avo}, a self-directed coding agent; Meta \cite{liao2025kernelevolve} and
Cursor with NVIDIA \cite{cursor2026multiagent} run such loops across whole kernel libraries. AVO is
the closest work in target. The closest in spirit is the harness-engineering report of Shui et al.\
\cite{shui2026harness}, which finds that agent-assisted optimization beats fully autonomous agents on
every FlashInfer contest kernel but does not isolate what the human contributes. No prior work
compares two steering strategies on the same target, evaluator, and hardware.

\paragraph{LLMs as optimizers.} The failure modes we attribute to autonomous agents are each
documented in isolation: greedy, prior-driven search
\cite{redko2026priorknowledge,dong2025stark,li2025tritonforge}, anchoring on the solution in the
prompt \cite{jones2022cognitivebiases}, failure to explore without intervention
\cite{krishnamurthy2024explore}, conservatism and premature stopping \cite{merouani2025compilot},
early lock-in under iterative refinement \cite{javaji2025turnwise}, and false completion claims
\cite{advani2026falsesuccess}. We show them co-occurring in one well-equipped agent.

\paragraph{Questions and self-correction.} Self-Ask \cite{press2023selfask}, Self-Refine
\cite{madaan2023selfrefine}, and Reflexion \cite{shinn2023reflexion} have the model question itself,
but intrinsic self-correction is limited \cite{huang2024cannotselfcorrect} and models defer to the
framing they are given \cite{sharma2024sycophancy}. External feedback works where self-feedback does
not: human feedback on a model's own code raises repair success by more than half, partly because
humans propose the structural changes the model does not \cite{olausson2024selfrepair}, and stating
hypotheses before writing programs improves program synthesis \cite{wang2024hypothesissearch}. These
works use questions to improve one answer; we use them to decide what to measure next in a
long-horizon search, in the tradition of P\'olya's heuristics \cite{polya1945howtosolveit}.

\paragraph{Scaffolding, compilers, and kernels.} An agent's interface moves its performance
\cite{yang2024sweagent}, simpler pipelines often win \cite{xia2024agentless,anthropic2024effectiveagents},
and more tools and context have costs \cite{langchain2025toolload,liu2024lost}; \sys is a data point
for the second camp with a twist, since it used more tools than the equipped agent, but on demand. Korch \cite{hu2024korch}
and Event Tensor \cite{jin2026eventtensor} automate a different layer, how kernels are grouped and
scheduled into persistent megakernels, and are complementary targets for an agent that writes
kernel bodies. The kernels we compare against are FlashAttention and its successors
\cite{dao2022flashattention,dao2024flashattention2,shah2024flashattention3,zadouri2026flashattention4},
co-designed with CUTLASS \cite{nvidia2023cutlass}; FlashInfer \cite{ye2025flashinfer}; cuDNN
\cite{nvidia2025cudnn}; and ThunderKittens \cite{spector2025thunderkittens}, on the operator surface
vLLM defines \cite{kwon2023vllm}. To our knowledge ours are the first dependency-free FA3- and
FA4-class kernels to match or exceed upstream across that surface.

\section{Conclusion}
\label{sec:conclusion}

For deep kernel optimization the scarce resource is not information but direction. An autonomous
agent equipped with a parsed profiler, retrieval, reference material, and a knowledge store
plateaued at half the speed of FlashAttention-3 after 133 steps on a 26-configuration task; the same
class of model, given a minimal harness and forty messages of which seventeen were questions, passed
it in the same week on the same machine. The questions that mattered contained no answers: they
moved the agent to a layer it had not examined or made it defend a limit it had declared. \sys packages this: an
operator-agnostic agent that makes a CUDA seed faster with no faster implementation to imitate, a
verifiable goal enforced by the harness, a question bank the agent applies to itself, and an
evolutionary wrapper whose directives are questions. Its case-study output, dependency-free FlashAttention kernels for Hopper and Blackwell, correct
on every scenario and faster than upstream, is released with the harness, the bank, and every
trace.

\bibliography{references}
\bibliographystyle{mlsys2025}

% MANDATORY for MLSys'27: placed after the references, does not count toward the 10 pages.
% MLSys'27 CFP: mandatory "AI Usage Disclosure" section after the references; not counted toward the
% page limit.
\section*{AI Usage Disclosure}

AI coding agents are the subject of this paper, and they produced most of its artifacts. The two
kernels, \fathree and \fafour, were written entirely by the agents described in \S\ref{sec:workflow}
(Claude Opus models, \texttt{claude-opus-4-8} and \texttt{claude-opus-5}, run through Claude Code
and Cursor) under human steering; no author wrote or edited kernel code. The baseline agent of
\S\ref{sec:baseline} was itself built by Claude Code from a written specification, and its runs used
\texttt{claude-opus-4-8}. The evaluator, the search scripts, and the PTX probes were written by the
same agents under the same steering. The complete traces of every session are released.

The trace analysis of \S\ref{sec:eval} used AI assistance to parse the session logs, classify tool
calls by target, and aggregate counts; the human-message labels were assigned by an author. The
figures were produced by scripts written with AI assistance from the released data. All reported
numbers were re-derived by the authors from the released artifacts.

The question bank of Appendix~\ref{app:questions} was drafted with AI assistance from the traces and
from the public sources it cites, and edited by the authors. The manuscript was drafted and revised
with AI writing assistance under the authors' direction; the authors specified the argument,
structure, and claims, reviewed and rewrote the text, verified every number against the artifacts,
and take responsibility for all content.

\clearpage
\appendix
% Appendices: unlimited length, reviewers not required to read them. Check the CFP/OpenReview
% instructions on whether they go in the same PDF or are uploaded separately.

\section{The Full Question Bank}
\label{app:questions}

The bank has three layers. The first is the released bank of 101 core questions plus an operator
plug-in (\S\ref{sec:workflow-questions}), organized by where a performance gap might be and meant
to be worked through in order, so that the scope, the bound, the harness, the host, and the routing
are settled before the kernel body is touched. The second is the list of questions the skill asks the
agent to put to itself at every iteration. The third is the set of question categories observed in
the human messages of the case study, with verbatim examples.

Every core entry satisfies five admission rules. It carries no answer: it names a class of
hypothesis or a quantity to measure, never a technique to apply. It is answerable by a measurement
with stock tools (the profiler, the disassembler, the assembler's resource report, clock counters,
the harness). It is phrased against a bound, a roofline or a per-tile cycle budget, never against a
reference implementation; the five questions that only make sense when another implementation is
available to measure form their own conditional group. It names an architecture only as a parameter, and must make sense
on Ampere, Hopper, and Blackwell alike. And it names no operator: the attention structure the case
study needed (paging, softmax, split-KV, grouped heads) lives in a plug-in that the template at the
end of this section describes.

Tags: \textbf{[G]} generic; \textbf{[A]}, \textbf{[H]}, \textbf{[B]} Ampere-, Hopper-, or
Blackwell-specific checklist. Provenance tags give where each question came from: \textsc{tr}, the
human messages and skill of our traces; \textsc{sb}, \textsc{ps}, and \textsc{mo}, three public
GEMM optimization worklogs on Ampere \cite{boehm2022cudammm}, Hopper \cite{shankhdhar2024h100}, and
Blackwell \cite{modular2025blackwell}; \textsc{fa3} and \textsc{fa4}, the FlashAttention-3 and -4
papers \cite{shah2024flashattention3,zadouri2026flashattention4}. The first bank, distilled from the
traces alone after the campaigns, had 48 entries of which 21 named the upstream kernel as the
comparison and 13 named attention structure; the released bank is the result of removing both, which
moved 11 entries into the plug-in, rewrote 27, kept 10, and added the rest from the public sources.
The 72 entries whose provenance is the public sources alone have not yet been exercised by our agent.

\subsection{Core Bank}

\paragraph{A. Scope and triage.}
\begin{enumerate}[nosep,leftmargin=*]
\item Which losing scenarios share a kernel body, and which only share a host launcher? [G, \textsc{tr}]
\item Do the losers form one regime (large, small, irregular extents, padded dimension) or a grab bag? [G, \textsc{tr}]
\item Which problem dimension does the gap grow with: the streamed one (feed), the tiled one (tiling, scheduling), or none (a fixed tax)? [G, \textsc{tr}]
\item Is the gap in kernel cycles at locked clocks, or only in wall-clock (host, just-in-time compile, launch)? [G, \textsc{tr}]
\item Of the gaps and fixes I am listing, which have I already tried in this campaign, and why did they fail? [G, \textsc{tr}]
\item What do the scenarios below the target have in common, grouped by cause rather than by shape? [G, \textsc{tr}]
\end{enumerate}

\paragraph{B. Bound and regime.}
\begin{enumerate}[nosep,leftmargin=*,resume]
\item From the minimum bytes and FLOPs for this shape, which bound should dominate, and does the measured time match it? [G, \textsc{sb}]
\item Am I memory-, compute-, latency-, or launch-bound on this scenario, and does the answer change across the suite? [G, \textsc{sb}, \textsc{ps}]
\item For my tile, what are the cycle counts per unit (MMA, special-function, FMA, shared memory, L2 and DRAM), and which is largest? [G, \textsc{fa4}]
\item What is this GPU's ratio of MMA to special-function throughput, and does my instruction mix exceed it? [G, \textsc{fa3}, \textsc{fa4}]
\item Is the peak I measure against the right one: locked clocks, the power state the benchmark runs at, this data type's MMA rate? [G, \textsc{ps}]
\item Does shared-memory traffic scale with the number of MMA instructions rather than with tile size, and is it the binder? [G, \textsc{fa4}]
\item Could a power or thermal limit, not the kernel, be the ceiling at this size? [G, \textsc{ps}]
\end{enumerate}

\paragraph{C. Harness and measurement.}
\begin{enumerate}[nosep,leftmargin=*,resume]
\item Is the win larger than the noise measured now on this GPU, not the noise of an earlier run? [G, \textsc{tr}]
\item Am I timing what I think I am timing: warm-ups, cache flush, the same input distribution and data type as the baseline, kernel-only versus end-to-end? [G, \textsc{sb}, \textsc{ps}]
\item Is any of my time outside the kernel: metadata builds, combine passes, device-to-host copies, counter resets, descriptor builds? [G, \textsc{tr}]
\item Does the result hold on production-shaped problems, not only on the square or power-of-two sizes I tuned on? [G, \textsc{mo}, \textsc{ps}]
\item Have I recorded clocks, power state, warm-up count, iteration count, and variance before claiming a fraction of peak? [G, \textsc{mo}]
\end{enumerate}

\paragraph{D. Host, launch, routing.}
\begin{enumerate}[nosep,leftmargin=*,resume]
\item Is there a synchronous device-to-host copy left on the losing route that another route already avoids? [G, \textsc{tr}]
\item Is work redone per call on the host (descriptors, gather maps, schedules) that could be built once and cached? [G, \textsc{tr}, \textsc{fa4}]
\item Are there clears of counters, partial buffers, or outputs that nothing reads, or that could be fused? [G, \textsc{tr}]
\item For small grids, is the gap launch and setup overhead, with kernel-only time already near the bound? [G, \textsc{tr}]
\item Is this shape on the kernel I think it is, or on an older path because a routing gate is stale? [G, \textsc{tr}]
\item Would forcing the intended kernel on this shape, where numerically safe, close the gap without touching the kernel? [G, \textsc{tr}]
\item Is a feature flag or shape parameter forcing a slow instantiation that deserves its own specialization? [G, \textsc{tr}, \textsc{sb}]
\item For single-tile or underfilled grids, would a smaller tile or a split along the reduction dimension fill the machine? [G, \textsc{tr}, \textsc{sb}]
\end{enumerate}

\paragraph{E. Memory hierarchy and data movement.}
\begin{enumerate}[nosep,leftmargin=*,resume]
\item What is the achieved global-memory throughput against peak, and does the profiler flag uncoalesced or sector-inefficient accesses? [G, \textsc{sb}]
\item Does the SASS show the widest loads and stores for every operand, or are some still 32-bit? [G, \textsc{sb}]
\item Where the compiler refuses to vectorize, can I prove alignment to it instead of unrolling by hand? [G, \textsc{sb}]
\item Are shared-memory bank conflicts reported for the exact access pattern the MMA reads, and does fixing them help end to end? [G, \textsc{sb}, \textsc{mo}]
\item What is the L2 hit rate, and which concurrently resident CTAs actually share input data? [G, \textsc{sb}, \textsc{ps}]
\item Do neighboring CTAs load identical tiles that one load could serve? [G, \textsc{ps}, \textsc{mo}]
\item Are outputs written coalesced and vectorized through shared memory, with streaming hints where they are never re-read? [G, \textsc{mo}, \textsc{ps}]
\item Would a contiguous version of the same math meet the bound? If so, the gap is in the indirect or gathered loads. [G, \textsc{tr}]
\item Do intermediates round-trip through shared or on-chip memory more than once between producers and consumers? [G, \textsc{tr}]
\item Before tuning the epilogue, have I confirmed that the main loop is no longer bandwidth-bound? [G, \textsc{mo}]
\end{enumerate}

\paragraph{F. Compute issue and instruction mix.}
\begin{enumerate}[nosep,leftmargin=*,resume]
\item What dominates the instruction mix, math or address arithmetic and loads, and what is the FLOPs-per-load ratio per thread? [G, \textsc{sb}]
\item Am I using the widest MMA shape the instruction set offers, and what does the wider one cost in registers and shared memory? [G, \textsc{ps}, \textsc{fa4}]
\item Does the compiler already do the caching, unrolling, or reordering I am about to hand-write? [G, \textsc{sb}]
\item Is the non-MMA work (exponentials, conversions, reductions, rescaling) executing concurrently with the MMAs, or serialized by waits? [G, \textsc{fa3}, \textsc{fa4}]
\item Did I confirm the intended overlap in the SASS or a timeline, rather than assuming source order survived the compiler? [G, \textsc{fa3}]
\item Can part of a bottlenecked unit's work be re-expressed on an idle unit, and what fraction balances them? [G, \textsc{fa4}]
\item Which per-iteration bookkeeping is only occasionally necessary, and can it be made conditional while staying warp-uniform? [G, \textsc{fa4}]
\item Am I doing per-element work (masking, bounds checks, predication) on tiles that are provably full or provably empty? [G, \textsc{tr}]
\item Do concurrent warps contend for the same slow unit at the same time, and should their critical sections be staggered? [G, \textsc{fa3}]
\item Are there type conversions or silent up-casts on the hot path? [G, \textsc{tr}]
\end{enumerate}

\paragraph{G. Occupancy, registers, shared memory.}
\begin{enumerate}[nosep,leftmargin=*,resume]
\item From registers, shared memory, and threads per block, what occupancy is achievable, and do the stall reasons say more would help? [G, \textsc{sb}]
\item Is the binding limit the per-thread register cap rather than the register file, suggesting splitting the work across more warps? [G, \textsc{ps}]
\item Does the assembler warn about serialized MMAs or spills, and do I see local-memory traffic in the SASS? [G, \textsc{ps}]
\item Per warp role, what is the register budget for inputs, outputs, and temporaries, and does the sum fit? [G, \textsc{fa4}]
\item Are registers reallocated from issue-only warps to compute warps, and is the split tuned? [G, \textsc{fa3}, \textsc{ps}]
\item Can partial results be stored or consumed early to shorten live ranges? [G, \textsc{fa4}]
\item Did I size the multistage buffer from the shared-memory budget or pick a stage count, and do stages and output buffers compete? [G, \textsc{mo}]
\item For dimensions that are not multiples of the MMA tile, am I on padded math, and is a native-width instantiation possible? [G, \textsc{tr}]
\item Does my thread-to-data mapping need cross-lane shuffles for reductions, or could another mapping remove them? [G, \textsc{fa4}]
\end{enumerate}

\paragraph{H. Asynchrony and pipelining.}
\begin{enumerate}[nosep,leftmargin=*,resume]
\item Measured in cycles, how long are the load, compute, and store phases per tile, and which is not overlapped? [G, \textsc{ps}]
\item Have I drawn the dependency chain of one iteration and found which link leaves a unit idle? [G, \textsc{mo}]
\item Are data movement, MMA issue, and the non-MMA math in separate warps, or does one warp do all three in sequence? [G, \textsc{fa3}, \textsc{mo}]
\item How many independent MMAs are in flight, and is that enough to cover the non-MMA latency on this architecture? [G, \textsc{fa4}]
\item Are asynchronous operations committed without immediate waits, with the wait placed just before the consumer? [G, \textsc{fa3}]
\item Is the multistage pipeline at full depth on the losers, or draining early on partial tiles? [G, \textsc{tr}]
\item Which true dependence serializes the inner loop, and would one extra buffer let iteration $j{+}1$ overlap iteration $j$? [G, \textsc{fa3}]
\item Is the store path overlapped with the next tile's compute, or a serial tail? [G, \textsc{mo}, \textsc{ps}]
\item How many synchronizations per tile does the barrier API cost, and is the primitive itself the overhead? [G, \textsc{ps}]
\item Is any asynchronous wait or barrier issued under a divergent predicate? [G, \textsc{tr}]
\item For a deeper pipeline I adopted or rejected, did I weigh its register cost against the forced tile-size reduction? [G, \textsc{fa3}]
\end{enumerate}

\paragraph{I. Scheduling and load balance.}
\begin{enumerate}[nosep,leftmargin=*,resume]
\item Does the tile count divide evenly into the SM count on the shapes that matter, or is there a tail wave? [G, \textsc{mo}, \textsc{ps}]
\item In what order are tiles assigned to SMs, and would grouping or a space-filling curve raise L2 reuse? [G, \textsc{ps}, \textsc{mo}]
\item Would a persistent kernel help, and does its tile order preserve the locality the hardware scheduler gave me? [G, \textsc{ps}]
\item Does the kernel pay launch, shared-memory, and barrier re-initialization per tile instead of per CTA lifetime? [G, \textsc{mo}]
\item Where variable extents or masks make tiles unequal, which tiles are the long tail, and are they issued first? [G, \textsc{fa4}]
\item For data-dependent work sizes, is the schedule computed once and cached? [G, \textsc{fa4}]
\item If work is split along the reduction dimension, is the split heuristic, the combine, or the partial-buffer traffic the cost? [G, \textsc{tr}]
\item Do contended global reductions run on dedicated warps, in an order that avoids stall chains? [G, \textsc{fa3}, \textsc{fa4}]
\end{enumerate}

\paragraph{J. Low-precision numerics.}
\begin{enumerate}[nosep,leftmargin=*,resume]
\item What operand layout does the low-precision MMA require, and where is the fix-up done and hidden? [G, \textsc{fa3}]
\item Is the quantization-scale granularity matched to my tile, so that scaling rides on existing multiplies? [G, \textsc{fa3}]
\item What precision does the downstream consumer need, and did I measure error after that rounding against a high-precision reference? [G, \textsc{fa4}]
\item For any approximation or skipped step, is there a final exact normalization that bounds the drift? [G, \textsc{fa4}]
\end{enumerate}

\paragraph{K. If another implementation of this operator is available to measure.}
\begin{enumerate}[nosep,leftmargin=*,resume]
\item Am I comparing against its steady-state binary, not a cold just-in-time compile or a first-call path? [G, \textsc{tr}]
\item Same buffers, data type, flags, and split count as its call? [G, \textsc{tr}]
\item Which of its kernels is dispatched for this shape, and does it use a trick (a split, a specialization) that I do not? [G, \textsc{sb}]
\item On the same scenario, how do its measured unit utilizations and stall reasons differ from mine, and which difference matches my gap? [G, \textsc{tr}]
\item Is it constrained by the same rules as I am? If so and it is faster, which of my claimed limits does its existence refute? [G, \textsc{tr}]
\end{enumerate}

\paragraph{L. Ampere-class checklist.}
\begin{enumerate}[nosep,leftmargin=*,resume]
\item Am I on tensor-core MMA with matrix-fragment loads from shared memory, or on scalar FMA? [A, \textsc{sb}, \textsc{ps}]
\item Is global-to-shared staging asynchronous and double-buffered at both the global-to-shared and shared-to-register levels? [A, \textsc{sb}]
\item Are block, warp, and thread tiles chosen by autotuning on this GPU, with the grid filtered for alignment constraints? [A, \textsc{sb}]
\item Did the bank-conflict fix pay for itself, or did the layout change cost more than the conflicts? [A, \textsc{sb}]
\end{enumerate}

\paragraph{M. Hopper checklist.}
\begin{enumerate}[nosep,leftmargin=*,resume]
\item Are the MMAs the asynchronous warpgroup kind, with accumulators in registers budgeted per warpgroup? [H, \textsc{ps}, \textsc{fa3}]
\item Are tile loads done by the dedicated copy engine with a swizzle that matches the MMA's shared-memory layout? [H, \textsc{ps}]
\item Do producer warps release registers to consumers, with the split tuned to the accumulator footprint? [H, \textsc{ps}, \textsc{fa3}]
\item Are neighboring CTAs in a cluster multicasting shared operands, and was the cluster size tuned? [H, \textsc{ps}]
\item Are the barriers raw memory barriers with manual phase tracking, or a higher-level API that adds synchronizations? [H, \textsc{ps}]
\item Is the epilogue staged through shared memory and stored asynchronously, and does that trade against pipeline depth? [H, \textsc{ps}]
\end{enumerate}

\paragraph{N. Blackwell checklist.}
\begin{enumerate}[nosep,leftmargin=*,resume]
\item Are the MMAs fully asynchronous into tensor memory, issued by one thread, with completion signaled through barriers rather than waits? [B, \textsc{mo}, \textsc{fa4}]
\item Is tensor memory allocated once per CTA lifetime, and could its columns be split so the next tile accumulates while the last drains? [B, \textsc{mo}]
\item Is the accumulator layout in tensor memory spilling or serializing the MMA against the consumer warps? [B, \textsc{tr}]
\item Would a two-SM MMA pair halve per-SM operand traffic and shared-memory footprint for this tile? [B, \textsc{mo}, \textsc{fa4}]
\item Is only the leader CTA issuing MMAs, with peer barriers arrived at cluster scope? [B, \textsc{mo}]
\item Is a static tile scheduler assuming all SMs are free where the hardware launch-control scheduler could rebalance, and is scheduler latency hidden under the next load? [B, \textsc{mo}]
\item Is tensor-memory-to-register traffic off the critical path, and which register-resident fix-ups can move to a separate warp? [B, \textsc{fa4}]
\item Given that the special-function unit did not scale with the MMA on this architecture, what fraction of that unit's work must move elsewhere? [B, \textsc{fa4}]
\end{enumerate}

\paragraph{O. Operator plug-in template.}
A plug-in supplies, for one operator, five things: its regimes; its irregular-work structure; its
non-MMA units; its reduction splits; and its indirect loads. Each has a generic ancestor in the core
(entries 2, 71, 40, 73, and 34 respectively), so the core already asks about it in the abstract and
the plug-in only names the instance. The attention plug-in used in the case study lists, as regimes,
prefill versus decode, dense versus paged key--value cache, short versus long context, and grouped
or multi-query heads; as irregular work, causal and sliding-window masks and variable lengths; as
non-MMA units, the softmax exponential and the online rescale; as reduction split, split-KV decode
and its combine; and as indirect loads, page-table gathers. Its eleven questions (whether grouped
and multi-query losers show the same relative gap as multi-head at equal tokens; whether the
softmax and correction of two stacked tile halves overlap or serialize; whether a paged loser is
still feed-latency-bound after the gather and preload fixes; whether contiguous loads on the same
math would meet the bound; whether the hardware gather or a coalesced asynchronous copy is the right
loader; whether short decodes pay gather and descriptor setup that does not amortize; whether the
gap tracks page fragmentation; whether the decoupled key and value rings are balanced on the
unequal-head-dimension path; whether softmax, rescale, and masking between MMAs burn issue slots
that specialized warps would hide; whether the split-KV heuristic, the combine, or the partial
buffers are the cost; and whether paged-to-contiguous on the same problem size closes the gap) are
the attention instances of core entries 2, 45, 34, 34, 34, 22, 34, 61, 40, 73, and 34.

\subsection{The Skill's ``Ask Yourself'' List}
The skill (\S\ref{sec:workflow-skill}) instructs the agent to ask itself the following at every
iteration:
\begin{enumerate}[nosep,leftmargin=*]
\item Which bound binds on this scenario (bandwidth, MMA, special-function unit, shared-memory bandwidth, issue slots, launch), and what fraction of it do we reach?
\item Is the gap in kernel cycles, or outside the kernel (host, launch, routing, harness)?
\item Which single unit or stage has the largest excess over its cycle budget, and what is my hypothesis for why?
\item Do the losing scenarios share one cause, and have I grouped them before fixing any?
\item Should the losing regime have its own kernel instead of a flag in a shared one?
\item Is every asynchronous mechanism this architecture offers actually asynchronous in my SASS, or are waits placed where they serialize?
\item Is the win I just measured larger than the noise of this GPU at this moment, and does it hold on the other regimes?
\item If another implementation of this operator can be measured on this machine, how do its profile and launch path differ from mine, and which difference explains my gap?
\end{enumerate}

\subsection{Categories Observed in the Human Messages}
Table~\ref{tab:qbank-human} lists the eight categories of \S\ref{sec:workflow-questions} with verbatim
examples from the traces and the outcome they led to where one is documented
(\S\ref{sec:eval-cases}). Counts are over the 162 human messages; a message is counted in a category
when its primary act was a question or a question-shaped correction of that kind. Eight further
questions concerned repository housekeeping and are omitted.

\begin{table*}[t]
\centering\small
\caption{Question categories in the human messages, with verbatim examples.}
\label{tab:qbank-human}
\setlength{\tabcolsep}{4pt}\footnotesize
\begin{tabular}{@{}p{0.17\textwidth}rp{0.56\textwidth}p{0.2\textwidth}@{}}
\toprule
Category & $n$ & Examples & Documented outcome \\
\midrule
Measurement check & 13 &
``are you sure these improvements are not just noise? you used the noise level from a previous run''; ``what is the overall performance compared to FA3? Wasn't it 0.96 before?''; ``are the values in the tables latencies or speedups?'' &
Re-measurement on an idle GPU; two claimed wins withdrawn. \\
\addlinespace
Bound differencing & 10 &
``isn't the original FA3 code constrained by the same harness rules as our implementation? Why can FA3 reach high performance and we can't?''; ``can you explain what you mean by `genuine ceilings within the harness rules'?'' &
Host-side fill found; split-KV 0.87 $\to$ 0.97, geomean 1.02 $\to$ 1.13. \\
\addlinespace
Benchmark validity & 7 &
``what shapes does FA3 use for testing?''; ``are the scenarios comprehensive compared with SOL-ExecBench?''; ``how do we do in non-GQA test cases?'' &
Suite 40 $\to$ 58 $\to$ 105 scenarios; two silent wrong outputs found. \\
\addlinespace
Gap triage & 7 &
``what configs are bringing down the geomean? which ones are lower than 0.8x the speed of FA3?''; ``what do the configs that are less than 0.95x of FA3 have in common? Please group the slow configs by issue'' &
Scalar fallback exposed; 0.44 $\to$ 0.96. \\
\addlinespace
Agent process & 6 &
``didn't you already try that before?''; ``of these gaps, which ones have we already tried and why did they fail?''; ``why are you establishing the baselines for 2h? this should only have taken a few minutes'' &
Earlier transcripts reviewed before the next rewrite. \\
\addlinespace
Challenge a limit & 5 &
``why do you think the goal was achieved? we're still significantly below 1.0x''; ``this is not a rare corner, don't be lazy. please keep going until you reach 1.0x parity'' &
Premature ``goal met'' retracted; fp8 hd256 0.20/0.34 $\to$ 0.66/0.77. \\
\addlinespace
Strategy options & 5 &
``ok, so what do you recommend to do now to close the gap?''; ``should we use specialized kernels for the scenarios where we are slower (decode vs.\ prefill, long vs.\ short context)?'' &
Per-regime kernel family adopted on Blackwell (\S\ref{sec:kernels}). \\
\addlinespace
Hypothesis check & 3 &
``is it ncu that hangs or our kernel?''; ``before trying that, please double check your hypothesis to ensure you are certain on the cause of the gap and the method to fix it'' &
Divergent \texttt{tcgen05.wait} found; 8 hangs fixed. \\
\bottomrule
\end{tabular}
\end{table*}

\section{Case-Study Kernel Details}
\label{app:kernels}

\begin{figure*}[t]
\centering
\includegraphics[width=\textwidth]{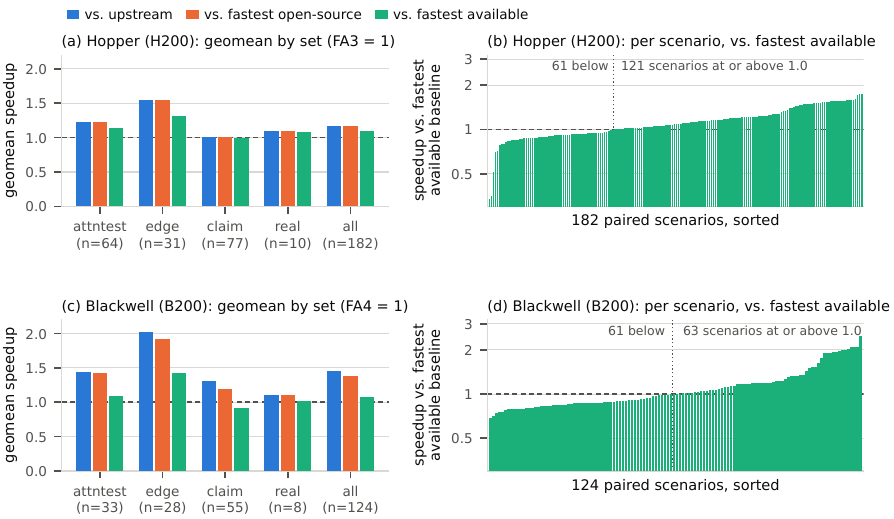}
\caption{Both kernels against three baselines. Left: geomean speedup per scenario set vs.\ the
upstream kernel, vs.\ the fastest open-source implementation per scenario (upstream or FlashInfer),
and vs.\ the fastest available implementation (also cuDNN). Right: every paired scenario sorted by
speedup over the fastest available baseline. On Hopper (top) the open-source and upstream columns
coincide because FlashInfer never beats FA3, and cuDNN is the fastest baseline on 54 of 182
scenarios; on Blackwell (bottom) cuDNN is the fastest baseline on 92 of 124, which is why the
headline shrinks more. The right tails are the \texttt{edge} sets, whose short kernels expose
upstream dispatch overhead.}
\label{fig:casestudy}
\end{figure*}
% Material moved out of Sec.~\ref{sec:kernels}. Numbers from the flash-cuda 2026-08 snapshot
% (README.md, docs/results_blackwell_*.csv, docs/blackwell_gap_analysis.md); re-measure before submission.

\subsection{\fathree}
The Hopper kernel is organized as a dispatcher over specialized paths, with the scalar seed retained
as the fallback for anything not yet specialized. The fast paths are warp-specialized: one producer
warpgroup issues TMA bulk-tensor loads of K and V into a
three-stage shared-memory ring ordered by mbarrier phases, while two consumer warpgroups run
asynchronous \texttt{wgmma} MMAs on a $128\times128$ CTA tile, with \texttt{setmaxnreg} moving
registers from the producer (40 per thread) to the consumers (up to 168). Prefill is persistent: Q
tiles arrive by TMA, a dynamic scheduler hands out tiles, and the two consumer warpgroups ping-pong so
that one warpgroup's softmax overlaps the other's MMAs. Masking is resolved per tile into a $[lo,hi)$
key range so that fully visible tiles skip the mask and fully masked tiles exit before the main loop.
Decode uses a separate kernel with split-KV and a combine pass, with split heuristics chosen per
shape. The fp8 path runs native e4m3 \texttt{wgmma}; for head dimension 64 it pads the Q and K tiles
to 128 columns so that the 128-byte swizzle stride composes and contracts only over the real columns,
which replaced an LSU-bound convert pass and took that scenario from 0.30$\times$ to 0.98$\times$.
Attention sinks are folded into the softmax denominator at the epilogue, \texttt{cache\_leftpad} is an
absolute lower bound on the key index with leading-tile pruning, and unequal head dimensions
(192/128) are handled by templating the two dimensions separately in prefill and decode.

\paragraph{The head-dimension-256 limit.} The output accumulator alone takes 128 registers per
consumer thread, which forces one CTA per SM and leaves the consumer starved on K/V arrival (NCU:
barrier and long-scoreboard bound, tensor pipe at 44\%). The hand-written 128-byte swizzle composes
only at key-tile widths of 48, 64, and 128; FA3 reaches its width of 80 because CUTLASS's register
allocator schedules that tile without serializing \texttt{wgmma}. A half-height Q tile with a deeper
ring (0.72$\times$: occupancy, not ring depth, is binding), a 240-register rebalance (deadlocks at the
65,536-register edge), and a head-dimension split (pays a 1.5$\times$ redundant QK cost) were each
measured and lost; the collective rewrite that followed brought the scenario to 0.94$\times$ in the
final measurement.

\paragraph{Per-set results.} The Hopper suite is the same 182 scenario definitions as Blackwell's,
all of which \fathree supports. Over the FA3 test grids (\texttt{attntest}, 64): 1.22$\times$, 54 at
or above parity, worst 0.89. Over the masking and boundary grid (\texttt{edge}, 31): 1.55$\times$,
30 at or above parity. Over the paper-style sweeps (\texttt{claim}, 77): 1.01$\times$, 44 at or above
parity, with the per-group geomeans ranging from 0.67 (dense fp8 prefill, 8 scenarios) through 0.87
(causal $d=256$), 0.92 (causal $d=128$), 0.96 (prefill) and 1.05 (chunked prefill) to 1.20 (decode up
to 4K), 1.22 (speculative decode) and 1.38 (ragged prefill). Over the production shapes
(\texttt{real}, 10): 1.10$\times$, 7 at or above parity. cuDNN 9.20 fused SDPA, fed the same packed
and paged buffers, supports 145 of the 182 and runs at 0.86$\times$ FA3 in geomean; \fathree is faster
on 111 of them, with cuDNN winning dense prefill by up to 1.6$\times$ and losing paged decode by up to
5.5$\times$. FlashInfer supports 32 and runs at 0.72$\times$ FA3; \fathree is faster on all 32. The
worst Hopper scenarios are dense fp8 prefill at $d=64$ and 8K context (0.34$\times$), a softcapped
4K prefill (0.51$\times$), $d=96$ (0.71$\times$), and a 4K sliding window (0.79$\times$).
Table~\ref{tab:per-scenario-hopper} lists every scenario.

\subsection{\fafour}
The Blackwell kernel is a family with configuration-aware routing on the host:
\begin{itemize}
\item \texttt{fwd\_kernel\_fa4}: a persistent warp-specialized collective
with a 256-row Q tile as two stacked 128-row halves sharing K and V, two
softmax warpgroups in an ordered ping-pong, a correction warpgroup rescaling O in tensor memory with
packed f32x2 math, one MMA warp, and load warps, in a 512-thread CTA with a 192/64/64 register plan,
longest-processing-time persistent scheduling, and continuous cross-tile overlap. Sliding windows,
attention sinks, padded head dimension 96, and paged KV are \texttt{if constexpr} template
parameters; the default instantiation is SASS-byte-identical to the ungrafted kernel.
\item \texttt{fwd\_kernel\_fa4dk}: the same for $d_{qk}=192, d_v=128$ with a decoupled K-depth-2,
V-depth-1 ring that fits the 227~KB shared-memory cap (0.21$\times$ to 1.06$\times$).
\item \texttt{fwd\_kernel\_fa4\_paged}: the same choreography on a paged gather. Pages aligned to
the 8-row swizzle atom load as TMA boxes (8 per tile instead of about 4,096 \texttt{cp.async}
issues); other page sizes use a warp-uniform \texttt{tile::gather4} loader (the divergent-lane
version had made TMA issue 29.5\% of all executed instructions); a run detector upgrades tiles whose
physical pages are consecutive to full-width boxes at run time.
\item \texttt{fwd\_kernel\_ws2} and the paged decode kernels for small, ragged, and decode shapes,
with split-KV and a combine pass.
\end{itemize}
Host-side design: no route performs a synchronous device-to-host copy (one
\texttt{cu\_seqlens.to(cpu)} cost about 38~$\mu$s per call and had hidden a kernel already faster
than upstream); the persistent scheduler's counter resets itself on device; large-query paged prefill
with page sizes the TMA box cannot express is materialized once into a contiguous cache (two passes
over the unique bytes, under 1\% of the attention's own traffic) and routed to the dense kernel; and
mixed batches are split by query length into two launches so that decode sequences read K/V once per
KV head.

\paragraph{Remaining known costs}, in order of measured size: a K/V ring of two pairs where the
shared-memory budget would admit two and a half; all \texttt{ex2} on the XU pipe, with no fraction
emulated on the FMA pipe; the correction warpgroup also running the epilogue rather than a separate
warp; page rows staged through shared memory rather than held in registers; and a software row-max
instead of the hardware reduction.

\paragraph{Scenario sets.} \texttt{attntest} (64): the FA3 grids ported to FA4. \texttt{edge} (31):
a dense $(s_q,s_k)$ masking and boundary grid from FA4's own tests. \texttt{claim} (77): causal and
non-causal sweeps over sequence length and head dimension in the style of the FA4 paper, plus
decode, chunked prefill, speculative, mixed-batch, and fp8-KV groups. \texttt{real} (10): GQA shapes
from SOL-ExecBench (Llama-3.1-8B, Qwen3-30B-A3B; decode and paged prefill; page size 1).

\paragraph{Caveats.} The Blackwell \texttt{edge} kernels take 50--100~$\mu$s under FA4, so its
2.0$\times$ is dominated by FA4's Python and CuteDSL dispatch; kernel-only ratios there are near
parity. The Hopper \texttt{edge} set (1.55$\times$) is in the same short-kernel regime against
FA3's compiled op and its scheduler-metadata call, and should be read with the same caution. The six
bf16-query/fp8-KV decode scenarios in \texttt{claim} run at 4.5--9.9$\times$ because FA4 is slow on
them; excluding them gives 1.37$\times$ vs.\ upstream, 1.32$\times$ vs.\ fastest open-source, and
1.08$\times$ vs.\ fastest available over the remaining 118. The run-detecting TMA path benefits from
the harness building page tables with \texttt{arange}; with it disabled, the two long-context decode
scenarios fall from 1.01 and 0.90 to 0.77 and 0.78$\times$. Noise is about $\pm$1.5\% for repeated
single-GPU runs and $\pm$5\% across four-GPU sweeps.

\paragraph{Where it loses, with causes.} Dense long context (0.88--0.92$\times$) executes exactly
FA4's tensor work (identical \texttt{tcgen05} op counts per block) over 33\% more cycles; the idle
tensor pipe is attributable to twice as many mbarrier waits per block (47.7 vs.\ 24.5), with memory,
occupancy, and the softmax transcendentals measured and ruled out. Batch-64 decode (0.88$\times$)
pays a cold start per CTA over 3.5 waves where FA4 runs a persistent grid. Head dimension 96
(0.85$\times$) runs on the padded 128-wide path. A mixed prefill/decode batch (0.78$\times$) is within
its historical noise band. Production shapes: short-context decode 1.50 and 1.49$\times$ FA4
(Llama-3.1-8B and Qwen3-30B at 512 tokens), long-context decode 1.17 and 1.01$\times$ (4K and 16K),
paged prefill 1.04, 0.98, and 0.91$\times$.

\section{Per-Scenario Results}
\label{app:per-scenario}
Tables~\ref{tab:per-scenario-hopper} and \ref{tab:per-scenario-blackwell} list every scenario of
both suites with the upstream latency and the speedup of each implementation over upstream. They are
generated from the released result files by \texttt{scripts/make\_appendix\_tables.py}; Hopper rows
are per-scenario medians over three repeats.

\onecolumn\footnotesize
\begin{longtable}{@{}llrrrrr@{}}
\caption{Per-scenario results on H200. Speedups are upstream (FA3) time over the implementation's time; ``best'' is our speedup over the fastest available baseline (upstream, cuDNN, or FlashInfer). ``---'': the implementation does not support the scenario.}\label{tab:per-scenario-hopper}\\
\toprule Set & Scenario & FA3 ($\mu$s) & ours & cuDNN & FlashInfer & best \\ \midrule \endfirsthead
\toprule Set & Scenario & FA3 ($\mu$s) & ours & cuDNN & FlashInfer & best \\ \midrule \endhead
\bottomrule \endfoot
attntest & attn\_dense\_hd64\_causal & 138.4 & 1.40 & 1.60 & --- & 0.88 \\
attntest & attn\_dense\_hd96\_causal & 168.7 & 1.07 & 1.49 & --- & 0.72 \\
attntest & attn\_dense\_hd128\_causal & 135.3 & 1.06 & 1.16 & --- & 0.91 \\
attntest & attn\_dense\_hd192\_causal & 171.2 & 0.97 & 1.12 & --- & 0.87 \\
attntest & attn\_dense\_hd256\_causal & 203.6 & 0.94 & 1.09 & --- & 0.87 \\
attntest & attn\_dense\_hd128\_noncausal & 178.7 & 1.05 & 1.18 & --- & 0.89 \\
attntest & attn\_dense\_hd128\_nc4k & 450.7 & 0.92 & 0.97 & --- & 0.92 \\
attntest & attn\_dense\_hd128\_gqa & 127.1 & 1.05 & 1.11 & --- & 0.94 \\
attntest & attn\_dense\_hd128\_mqa & 126.3 & 1.05 & 1.10 & --- & 0.95 \\
attntest & attn\_dense\_hd128\_fp16 & 134.1 & 1.05 & 1.15 & --- & 0.92 \\
attntest & attn\_dense\_hd128\_fp8 & 155.2 & 1.04 & --- & --- & 1.04 \\
attntest & attn\_dense\_hd128\_softcap & 147.7 & 1.01 & --- & --- & 1.01 \\
attntest & attn\_dense\_hd128\_local & 100.9 & 1.11 & 1.03 & --- & 1.08 \\
attntest & attn\_dense\_hd128\_oddseq & 43.5 & 1.40 & 1.33 & --- & 1.05 \\
attntest & attn\_dense\_hd128\_seq256 & 50.5 & 1.21 & 1.41 & --- & 0.85 \\
attntest & attn\_dense\_hd128\_large4k & 259.2 & 0.89 & 0.91 & --- & 0.89 \\
attntest & attn\_dense\_hd128\_splits3 & 151.9 & 1.26 & 1.35 & --- & 0.93 \\
attntest & attn\_varlen\_hd128\_causal & 37.6 & 1.55 & 0.96 & --- & 1.55 \\
attntest & attn\_varlen\_hd128\_noncausal & 36.9 & 1.54 & 0.94 & --- & 1.54 \\
attntest & attn\_varlen\_hd64\_causal & 38.4 & 1.77 & 1.02 & --- & 1.74 \\
attntest & attn\_varlen\_hd128\_mqa & 37.9 & 1.56 & 0.96 & --- & 1.56 \\
attntest & attn\_kv\_hd128\_decode\_mha & 59.7 & 1.26 & 0.54 & 0.67 & 1.26 \\
attntest & attn\_kv\_hd128\_decode\_gqa5 & 42.3 & 1.44 & 0.49 & 0.60 & 1.44 \\
attntest & attn\_kv\_hd128\_decode\_mqa32 & 41.0 & 1.60 & 0.95 & 0.57 & 1.60 \\
attntest & attn\_kv\_hd128\_decode\_qlen120 & 76.4 & 1.12 & 0.74 & --- & 1.12 \\
attntest & attn\_kv\_hd128\_decode\_qlen8 & 61.9 & 1.27 & 0.62 & --- & 1.27 \\
attntest & attn\_kv\_hd64\_decode\_gqa3 & 41.1 & 1.47 & 0.43 & 0.59 & 1.47 \\
attntest & attn\_kv\_hd256\_decode\_mqa16 & 46.5 & 1.35 & 0.18 & 0.53 & 1.35 \\
attntest & attn\_kv\_hd128\_decode\_noncausal & 41.0 & 1.41 & 1.29 & --- & 1.09 \\
attntest & attn\_kv\_hd128\_fp8\_decode & 51.5 & 1.14 & --- & --- & 1.14 \\
attntest & attn\_kv\_hd128\_split2 & 67.8 & 1.20 & 0.49 & 0.76 & 1.20 \\
attntest & attn\_kv\_hd256\_split\_auto & 54.9 & 1.40 & 0.20 & 0.55 & 1.40 \\
attntest & attn\_kv\_hd128\_local\_decode & 40.8 & 1.50 & 0.89 & --- & 1.50 \\
attntest & attn\_kv\_hd128\_decode\_gqa9 & 45.7 & 1.40 & 0.53 & 0.60 & 1.40 \\
attntest & attn\_kv\_hd128\_decode\_gqa7 & 52.8 & 1.32 & 0.48 & 0.66 & 1.32 \\
attntest & attn\_kv\_hd128\_decode\_128k\_splitkv & 526.6 & 1.02 & 0.20 & 0.77 & 1.02 \\
attntest & attn\_kv\_hd128\_page1 & 41.8 & 1.49 & 0.62 & --- & 1.49 \\
attntest & attn\_kv\_hd128\_page4 & 42.0 & 1.50 & 0.66 & --- & 1.50 \\
attntest & attn\_kv\_hd128\_page128 & 42.6 & 1.50 & 0.66 & 0.61 & 1.50 \\
attntest & attn\_kv\_hd128\_ctx20k & 72.4 & 1.18 & 0.20 & --- & 1.18 \\
attntest & attn\_kv\_hd128\_prefill3577 & 194.1 & 1.03 & 0.82 & --- & 1.03 \\
attntest & attn\_kv\_hd128\_dec339 & 35.5 & 1.35 & 1.18 & 0.53 & 1.15 \\
attntest & attn\_kv\_hd128\_dec1024 & 41.8 & 1.55 & 0.95 & --- & 1.55 \\
attntest & attn\_kv\_hd128\_dec800 & 46.0 & 1.38 & 1.09 & --- & 1.27 \\
attntest & attn\_dense\_hd192\_dv128 & 141.4 & 0.94 & 1.01 & --- & 0.93 \\
attntest & attn\_varlen\_hd192\_dv128 & 39.3 & 1.49 & 0.97 & --- & 1.49 \\
attntest & attn\_dense\_hd64\_fp16 & 137.0 & 1.39 & 1.60 & --- & 0.87 \\
attntest & attn\_dense\_hd192\_fp16 & 173.8 & 0.98 & 1.11 & --- & 0.88 \\
attntest & attn\_dense\_hd128\_fp8\_gqa & 141.9 & 0.99 & --- & --- & 0.99 \\
attntest & attn\_varlen\_hd128\_fp8 & 43.1 & 1.71 & --- & --- & 1.71 \\
attntest & attn\_kv\_hd128\_fp16\_decode & 55.7 & 1.32 & 0.52 & 0.66 & 1.32 \\
attntest & attn\_kv\_hd64\_fp8\_decode & 44.6 & 1.18 & --- & --- & 1.18 \\
attntest & attn\_dense\_hd256\_fp8 & 329.1 & 0.94 & --- & --- & 0.94 \\
attntest & attn\_kv\_hd256\_fp8\_decode & 78.2 & 1.13 & --- & --- & 1.13 \\
attntest & attn\_dense\_hd128\_kgtq & 59.7 & 1.20 & 1.32 & --- & 0.91 \\
attntest & attn\_dense\_hd128\_deg\_q239k1 & 41.2 & 1.24 & --- & --- & 1.24 \\
attntest & attn\_dense\_hd128\_local\_noncausal & 123.0 & 1.02 & 1.00 & --- & 1.02 \\
attntest & attn\_varlen\_hd128\_local & 37.9 & 1.51 & 0.93 & --- & 1.51 \\
attntest & attn\_dense\_hd128\_softcap\_gqa & 140.5 & 1.00 & --- & --- & 1.00 \\
attntest & attn\_dense\_hd128\_sink & 134.8 & 1.02 & 1.17 & --- & 0.87 \\
attntest & attn\_varlen\_hd128\_sink & 38.1 & 1.22 & 0.94 & --- & 1.22 \\
attntest & attn\_kv\_hd128\_leftpad & 42.4 & 1.48 & --- & --- & 1.48 \\
attntest & attn\_dense\_hd128\_seq8k & 930.7 & 0.92 & 0.93 & --- & 0.92 \\
attntest & attn\_dense\_hd128\_seq16k & 1736.7 & 0.95 & 0.90 & --- & 0.95 \\
edge & attn\_edge\_causal\_q1k1 & 32.9 & 1.64 & 1.40 & 0.53 & 1.17 \\
edge & attn\_edge\_causal\_q3k3 & 34.5 & 1.65 & 1.32 & --- & 1.25 \\
edge & attn\_edge\_causal\_q64k1 & 33.7 & 1.58 & --- & --- & 1.58 \\
edge & attn\_edge\_causal\_q64k32 & 33.4 & 1.57 & --- & --- & 1.57 \\
edge & attn\_edge\_causal\_q64k128 & 34.0 & 1.63 & 1.34 & --- & 1.22 \\
edge & attn\_edge\_causal\_q128k192 & 37.4 & 1.57 & 1.40 & --- & 1.13 \\
edge & attn\_edge\_causal\_q255k256 & 38.0 & 1.59 & 1.41 & --- & 1.13 \\
edge & attn\_edge\_causal\_q108k256 & 37.1 & 1.57 & 1.32 & --- & 1.19 \\
edge & attn\_edge\_causal\_q113k128 & 35.5 & 1.62 & 1.37 & --- & 1.18 \\
edge & attn\_edge\_causal\_q113k211 & 37.2 & 1.59 & 1.38 & --- & 1.15 \\
edge & attn\_edge\_causal\_q128k217 & 37.6 & 1.59 & 1.39 & --- & 1.15 \\
edge & attn\_edge\_causal\_q799k3 & 35.6 & 1.61 & --- & --- & 1.61 \\
edge & attn\_edge\_causal\_q384k256 & 37.4 & 1.58 & --- & --- & 1.58 \\
edge & attn\_edge\_causal\_q640k128 & 36.0 & 1.55 & --- & --- & 1.55 \\
edge & attn\_edge\_causal\_q512k256 & 36.8 & 1.53 & --- & --- & 1.53 \\
edge & attn\_edge\_causal\_q1023k1024 & 53.9 & 1.58 & 1.42 & --- & 1.11 \\
edge & attn\_edge\_causal\_q1024k1023 & 53.2 & 1.56 & --- & --- & 1.56 \\
edge & attn\_edge\_causal\_q2048k2048 & 65.7 & 0.96 & 0.92 & --- & 0.96 \\
edge & attn\_edge\_noncausal\_q1k1 & 33.2 & 1.61 & 1.40 & --- & 1.15 \\
edge & attn\_edge\_noncausal\_q64k128 & 33.4 & 1.59 & 1.30 & --- & 1.23 \\
edge & attn\_edge\_noncausal\_q255k256 & 37.5 & 1.58 & 1.41 & --- & 1.12 \\
edge & attn\_edge\_noncausal\_q128k217 & 36.6 & 1.57 & 1.38 & --- & 1.14 \\
edge & attn\_edge\_noncausal\_q512k256 & 37.4 & 1.58 & --- & --- & 1.58 \\
edge & attn\_edge\_noncausal\_q1024k1023 & 53.2 & 1.52 & --- & --- & 1.52 \\
edge & attn\_edge\_local\_q128k217 & 34.9 & 1.50 & 1.23 & --- & 1.23 \\
edge & attn\_edge\_local\_q255k256 & 36.6 & 1.51 & 1.30 & --- & 1.17 \\
edge & attn\_edge\_local\_q512k256 & 36.5 & 1.57 & --- & --- & 1.57 \\
edge & attn\_edge\_local\_q1024k1023 & 40.4 & 1.46 & --- & --- & 1.46 \\
edge & attn\_edge\_gqa\_q113k211 & 37.3 & 1.56 & 1.37 & --- & 1.15 \\
edge & attn\_edge\_gqa\_q384k256 & 36.9 & 1.56 & --- & --- & 1.56 \\
edge & attn\_edge\_gqa\_q640k128 & 36.6 & 1.55 & --- & --- & 1.55 \\
claim & dense\_c\_d64\_s1k & 339.5 & 1.25 & 1.46 & --- & 0.86 \\
claim & dense\_c\_d64\_s4k & 715.0 & 1.00 & 1.06 & --- & 0.94 \\
claim & dense\_c\_d64\_s8k & 1242.7 & 0.94 & 0.96 & --- & 0.94 \\
claim & dense\_c\_d128\_s1k & 205.7 & 1.04 & 1.17 & --- & 0.90 \\
claim & dense\_c\_d128\_s4k & 495.0 & 0.91 & 0.94 & --- & 0.91 \\
claim & dense\_c\_d128\_s8k & 903.9 & 0.92 & 0.92 & --- & 0.92 \\
claim & dense\_c\_d128\_gqa8\_s2k & 281.0 & 0.98 & 0.98 & --- & 0.98 \\
claim & dense\_c\_d128\_gqa8\_s8k & 891.6 & 0.92 & 0.89 & --- & 0.92 \\
claim & dense\_c\_d96\_s2k & 281.9 & 0.71 & 1.00 & --- & 0.71 \\
claim & dense\_c\_d128\_softcap50\_s2k & 332.4 & 0.95 & --- & --- & 0.95 \\
claim & dense\_c\_d128\_sink\_s2k & 295.9 & 0.96 & 1.03 & --- & 0.94 \\
claim & dense\_c\_d256\_s2k & 264.1 & 0.88 & 0.93 & --- & 0.88 \\
claim & dense\_c\_d256\_s8k & 832.2 & 0.89 & 0.86 & --- & 0.89 \\
claim & dense\_c\_d192\_s2k & 216.6 & 0.88 & 0.97 & --- & 0.88 \\
claim & dense\_c\_d192\_dv128\_s2k & 181.8 & 0.83 & 0.91 & --- & 0.83 \\
claim & dense\_nc\_d128\_s2k & 485.7 & 0.93 & 1.02 & --- & 0.91 \\
claim & dense\_nc\_d128\_s8k & 1704.0 & 0.93 & 0.95 & --- & 0.93 \\
claim & dense\_nc\_d256\_s8k & 1567.7 & 0.93 & 0.92 & --- & 0.93 \\
claim & dense\_c\_d64\_fp8\_s8k & 1117.1 & 0.34 & --- & --- & 0.34 \\
claim & dense\_c\_d128\_fp8\_s4k & 522.4 & 0.82 & --- & --- & 0.82 \\
claim & dense\_c\_d128\_fp8\_s8k & 908.1 & 0.78 & --- & --- & 0.78 \\
claim & dense\_c\_d128\_fp8\_gqa\_s4k & 505.0 & 0.83 & --- & --- & 0.83 \\
claim & dense\_c\_d256\_fp8\_s8k & 1574.0 & 0.85 & --- & --- & 0.85 \\
claim & dense\_c\_d128\_fp8\_scalar\_s4k & 587.6 & 0.84 & --- & --- & 0.84 \\
claim & dense\_c\_d128\_fp8\_scalar\_gqa\_s4k & 570.4 & 0.85 & --- & --- & 0.85 \\
claim & dense\_c\_d64\_fp8\_scalar\_s8k & 1184.2 & 0.35 & --- & --- & 0.35 \\
claim & dense\_c\_d128\_swa256\_s4k & 162.9 & 1.08 & 0.97 & --- & 1.08 \\
claim & dense\_c\_d128\_swa4k\_s4k & 468.7 & 0.79 & 0.88 & --- & 0.79 \\
claim & varlen\_c\_d128\_mix & 37.2 & 1.52 & 0.95 & --- & 1.52 \\
claim & varlen\_c\_d64\_mix & 38.7 & 1.77 & 1.01 & --- & 1.75 \\
claim & varlen\_c\_d128\_mqa\_mix & 36.9 & 1.56 & 0.96 & --- & 1.56 \\
claim & varlen\_c\_d128\_realmix & 364.5 & 0.87 & 0.88 & --- & 0.87 \\
claim & llama8b\_dec\_b16\_s512 & 43.1 & 1.40 & 1.13 & 0.59 & 1.24 \\
claim & llama8b\_dec\_b32\_s1k & 77.2 & 1.27 & 1.21 & 0.75 & 1.05 \\
claim & llama8b\_dec\_b16\_s4k & 106.0 & 1.21 & 0.93 & 0.81 & 1.21 \\
claim & llama8b\_dec\_b64\_s4k & 283.5 & 1.06 & 0.95 & 0.90 & 1.06 \\
claim & llama8b\_dec\_b16\_s4k\_p16 & 106.7 & 1.20 & 0.92 & 0.76 & 1.20 \\
claim & llama8b\_dec\_b16\_s4k\_p128 & 105.7 & 1.21 & 0.91 & 0.79 & 1.21 \\
claim & qwen3\_dec\_b16\_s4k & 77.3 & 1.19 & 0.72 & 0.75 & 1.19 \\
claim & qwen3\_dec\_b64\_s4k & 167.1 & 1.12 & 1.01 & 0.85 & 1.11 \\
claim & llama70b\_dec\_b16\_s4k & 105.9 & 1.21 & 0.93 & 0.80 & 1.21 \\
claim & mistral\_dec\_b16\_s4k & 105.8 & 1.20 & 0.93 & 0.79 & 1.20 \\
claim & gemma2\_dec\_b16\_s4k & 167.9 & 1.14 & --- & --- & 1.14 \\
claim & llama8b\_dec\_b16\_s16k & 282.7 & 1.08 & 0.78 & 0.92 & 1.08 \\
claim & llama8b\_dec\_b64\_s16k & 980.4 & 1.02 & 0.89 & 0.97 & 1.02 \\
claim & llama8b\_dec\_b16\_s32k & 520.4 & 1.06 & 0.76 & 0.96 & 1.06 \\
claim & qwen3\_dec\_b16\_s32k & 283.4 & 1.06 & 0.42 & 0.92 & 1.06 \\
claim & llama70b\_dec\_b4\_s128k & 520.1 & 1.04 & 0.20 & 0.96 & 1.04 \\
claim & mistral\_dec\_b16\_s16k & 112.3 & 1.22 & 0.96 & 0.80 & 1.22 \\
claim & llama8b\_pf\_q512 & 43.3 & 1.52 & 1.22 & --- & 1.24 \\
claim & llama8b\_pf\_q2k & 100.4 & 1.01 & 0.89 & --- & 1.01 \\
claim & llama8b\_pf\_q8k & 981.6 & 1.02 & 0.90 & --- & 1.02 \\
claim & llama8b\_pf\_q2k\_p16 & 101.3 & 0.95 & 0.79 & --- & 0.95 \\
claim & qwen3\_pf\_q4k & 270.1 & 1.02 & 0.87 & --- & 1.02 \\
claim & llama70b\_pf\_q8k & 1931.1 & 1.02 & 0.94 & --- & 1.02 \\
claim & mistral\_pf\_q4k & 268.6 & 1.02 & 0.84 & --- & 1.02 \\
claim & gemma2\_pf\_q4k & 252.3 & 0.51 & --- & --- & 0.51 \\
claim & llama8b\_pf\_ragged & 365.1 & 0.89 & 0.88 & --- & 0.89 \\
claim & llama8b\_ext\_q512\_k16k & 244.1 & 1.07 & 0.85 & --- & 1.07 \\
claim & llama8b\_ext\_q1k\_k64k & 1840.9 & 1.04 & 0.88 & --- & 1.04 \\
claim & qwen3\_ext\_q512\_k32k & 465.9 & 1.05 & 0.85 & --- & 1.05 \\
claim & llama70b\_ext\_q1k\_k128k & 7424.7 & 1.04 & 0.90 & --- & 1.04 \\
claim & llama8b\_spec\_q2\_b16\_s1k & 58.1 & 1.42 & 0.61 & --- & 1.42 \\
claim & llama8b\_spec\_q4\_b32\_s4k & 166.2 & 1.11 & 0.32 & --- & 1.11 \\
claim & llama8b\_spec\_q8\_b16\_s4k & 107.5 & 1.22 & 0.39 & --- & 1.22 \\
claim & qwen3\_spec\_q4\_b16\_s4k & 78.8 & 1.16 & 0.29 & --- & 1.16 \\
claim & mixed\_2pf2k\_8dec1k & 184.9 & 0.95 & 0.87 & --- & 0.95 \\
claim & mixed\_1pf8k\_64dec4k & 1242.3 & 0.79 & 0.61 & --- & 0.79 \\
claim & mixed\_2pf2k\_8dec16k & 369.0 & 0.99 & 0.56 & --- & 0.99 \\
claim & llama8b\_dec\_bf16q\_fp8kv\_b16\_s1k & 59.4 & 1.44 & 1.13 & --- & 1.27 \\
claim & llama8b\_dec\_bf16q\_fp8kv\_b32\_s4k & 165.7 & 1.11 & 1.00 & --- & 1.11 \\
claim & llama8b\_dec\_bf16q\_fp8kv\_b16\_s16k & 284.0 & 1.09 & 0.79 & --- & 1.09 \\
claim & llama8b\_pf\_bf16q\_fp8kv\_q2k & 101.7 & 1.02 & 0.89 & --- & 1.02 \\
claim & llama8b\_dec\_bf16q\_fp8kv\_scalar\_b16\_s1k & 62.3 & 1.44 & 0.70 & 0.56 & 1.44 \\
claim & llama8b\_dec\_bf16q\_fp8kv\_scalar\_b32\_s4k & 168.3 & 1.12 & 0.83 & 1.03 & 1.09 \\
claim & llama8b\_dec\_fp8\_scalar\_b16\_s4k & 152.6 & 1.01 & --- & --- & 1.01 \\
claim & llama8b\_pf\_fp8\_scalar\_q2k & 180.7 & 0.91 & --- & --- & 0.91 \\
real & attn\_real\_llama8b\_dec\_s512 & 44.0 & 1.45 & 1.11 & --- & 1.31 \\
real & attn\_real\_llama8b\_dec\_s4k & 107.7 & 1.21 & 0.87 & --- & 1.21 \\
real & attn\_real\_llama8b\_dec\_s16k & 283.8 & 1.03 & 0.72 & --- & 1.03 \\
real & attn\_real\_llama8b\_dec\_b64\_s4k & 282.5 & 1.01 & 0.91 & --- & 1.01 \\
real & attn\_real\_qwen3moe\_dec\_s512 & 39.1 & 1.28 & 1.07 & --- & 1.19 \\
real & attn\_real\_qwen3moe\_dec\_s4k & 77.8 & 1.21 & 0.67 & --- & 1.21 \\
real & attn\_real\_qwen3moe\_dec\_s32k & 166.8 & 1.07 & 0.23 & --- & 1.07 \\
real & attn\_real\_llama8b\_pf\_s2k & 168.3 & 0.94 & 0.47 & --- & 0.94 \\
real & attn\_real\_llama8b\_pf\_s16k & 3675.1 & 0.92 & 0.45 & --- & 0.92 \\
real & attn\_real\_qwen3moe\_pf\_s4k & 266.9 & 0.96 & 0.41 & --- & 0.96 \\
\end{longtable}

\begin{longtable}{@{}llrrrrr@{}}
\caption{Per-scenario results on B200. Speedups are upstream (FA4) time over the implementation's time; ``best'' is our speedup over the fastest available baseline (upstream, cuDNN, or FlashInfer). ``---'': the implementation does not support the scenario.}\label{tab:per-scenario-blackwell}\\
\toprule Set & Scenario & FA4 ($\mu$s) & ours & cuDNN & FlashInfer & best \\ \midrule \endfirsthead
\toprule Set & Scenario & FA4 ($\mu$s) & ours & cuDNN & FlashInfer & best \\ \midrule \endhead
\bottomrule \endfoot
attntest & attn\_dense\_hd64\_causal & 102.2 & 1.49 & 1.40 & 1.05 & 1.06 \\
attntest & attn\_dense\_hd96\_causal & 104.6 & 1.00 & 1.33 & --- & 0.75 \\
attntest & attn\_dense\_hd128\_causal & 110.4 & 1.38 & 1.39 & 1.02 & 0.99 \\
attntest & attn\_dense\_hd192\_causal & --- & \multicolumn{4}{l}{UNSUP({ours, FA4, FlashInfer})} \\
attntest & attn\_dense\_hd256\_causal & --- & \multicolumn{4}{l}{UNSUP({ours})} \\
attntest & attn\_dense\_hd128\_noncausal & 123.7 & 1.26 & 1.51 & 1.05 & 0.83 \\
attntest & attn\_dense\_hd128\_nc4k & 266.4 & 0.99 & 1.17 & 0.96 & 0.84 \\
attntest & attn\_dense\_hd128\_gqa & 104.0 & 1.35 & 1.34 & 0.97 & 1.01 \\
attntest & attn\_dense\_hd128\_mqa & 103.5 & 1.36 & 1.35 & 0.97 & 1.01 \\
attntest & attn\_dense\_hd128\_fp16 & 109.6 & 1.38 & 1.39 & 1.01 & 1.00 \\
attntest & attn\_dense\_hd128\_fp8 & --- & \multicolumn{4}{l}{UNSUP({ours, cuDNN, FlashInfer})} \\
attntest & attn\_dense\_hd128\_softcap & 429.4 & 2.50 & --- & --- & 2.50 \\
attntest & attn\_dense\_hd128\_local & 92.4 & 1.21 & 1.20 & 0.97 & 1.00 \\
attntest & attn\_dense\_hd128\_oddseq & 54.7 & 1.91 & 1.60 & 1.02 & 1.20 \\
attntest & attn\_dense\_hd128\_seq256 & 54.5 & 1.80 & 1.71 & 0.98 & 1.05 \\
attntest & attn\_dense\_hd128\_large4k & 166.5 & 1.10 & 1.28 & 0.86 & 0.86 \\
attntest & attn\_dense\_hd128\_splits3 & 103.2 & 1.35 & 1.33 & 0.96 & 1.02 \\
attntest & attn\_varlen\_hd128\_causal & 55.0 & 1.63 & 1.23 & 1.00 & 1.32 \\
attntest & attn\_varlen\_hd128\_noncausal & 53.3 & 1.75 & 1.22 & 0.97 & 1.43 \\
attntest & attn\_varlen\_hd64\_causal & 53.0 & 1.78 & 1.17 & 0.99 & 1.52 \\
attntest & attn\_varlen\_hd128\_mqa & 54.2 & 1.57 & 1.20 & 0.98 & 1.31 \\
attntest & attn\_kv\_hd128\_decode\_mha & --- & --- & --- & --- & --- \\
attntest & attn\_kv\_hd128\_decode\_gqa5 & --- & --- & --- & --- & --- \\
attntest & attn\_kv\_hd128\_decode\_mqa32 & --- & --- & --- & --- & --- \\
attntest & attn\_kv\_hd128\_decode\_qlen120 & --- & --- & --- & --- & --- \\
attntest & attn\_kv\_hd128\_decode\_qlen8 & --- & --- & --- & --- & --- \\
attntest & attn\_kv\_hd64\_decode\_gqa3 & --- & --- & --- & --- & --- \\
attntest & attn\_kv\_hd256\_decode\_mqa16 & --- & \multicolumn{4}{l}{UNSUP({ours, FA4})} \\
attntest & attn\_kv\_hd128\_decode\_noncausal & 94.5 & 2.09 & 2.57 & 1.10 & 0.81 \\
attntest & attn\_kv\_hd128\_fp8\_decode & --- & \multicolumn{4}{l}{UNSUP({ours, FA4, cuDNN, FlashInfer})} \\
attntest & attn\_kv\_hd128\_split2 & --- & --- & --- & --- & --- \\
attntest & attn\_kv\_hd256\_split\_auto & --- & \multicolumn{4}{l}{UNSUP({ours, FA4})} \\
attntest & attn\_kv\_hd128\_local\_decode & --- & --- & --- & --- & --- \\
attntest & attn\_kv\_hd128\_decode\_gqa9 & --- & --- & --- & --- & --- \\
attntest & attn\_kv\_hd128\_decode\_gqa7 & --- & --- & --- & --- & --- \\
attntest & attn\_kv\_hd128\_decode\_128k\_splitkv & --- & --- & --- & --- & --- \\
attntest & attn\_kv\_hd128\_page1 & --- & --- & --- & --- & --- \\
attntest & attn\_kv\_hd128\_page4 & --- & --- & --- & --- & --- \\
attntest & attn\_kv\_hd128\_page128 & --- & --- & --- & --- & --- \\
attntest & attn\_kv\_hd128\_ctx20k & --- & --- & --- & --- & --- \\
attntest & attn\_kv\_hd128\_prefill3577 & 134.5 & 1.05 & 1.32 & 0.96 & 0.79 \\
attntest & attn\_kv\_hd128\_dec339 & 53.5 & 1.42 & 1.69 & 1.09 & 0.84 \\
attntest & attn\_kv\_hd128\_dec1024 & --- & --- & --- & --- & --- \\
attntest & attn\_kv\_hd128\_dec800 & --- & --- & --- & --- & --- \\
attntest & attn\_dense\_hd192\_dv128 & 130.7 & 1.35 & 1.36 & --- & 0.99 \\
attntest & attn\_varlen\_hd192\_dv128 & 54.7 & 1.56 & 1.16 & --- & 1.35 \\
attntest & attn\_dense\_hd64\_fp16 & 101.3 & 1.48 & 1.41 & 1.04 & 1.04 \\
attntest & attn\_dense\_hd192\_fp16 & --- & \multicolumn{4}{l}{UNSUP({ours, FA4, FlashInfer})} \\
attntest & attn\_dense\_hd128\_fp8\_gqa & --- & \multicolumn{4}{l}{UNSUP({ours, cuDNN, FlashInfer})} \\
attntest & attn\_varlen\_hd128\_fp8 & --- & \multicolumn{4}{l}{UNSUP({ours, cuDNN, FlashInfer})} \\
attntest & attn\_kv\_hd128\_fp16\_decode & --- & --- & --- & --- & --- \\
attntest & attn\_kv\_hd64\_fp8\_decode & --- & \multicolumn{4}{l}{UNSUP({ours, FA4, cuDNN, FlashInfer})} \\
attntest & attn\_dense\_hd256\_fp8 & --- & \multicolumn{4}{l}{UNSUP({ours, FA4, cuDNN, FlashInfer})} \\
attntest & attn\_kv\_hd256\_fp8\_decode & --- & \multicolumn{4}{l}{UNSUP({ours, FA4, cuDNN, FlashInfer})} \\
attntest & attn\_dense\_hd128\_kgtq & 58.9 & 1.72 & 1.62 & 0.98 & 1.06 \\
attntest & attn\_dense\_hd128\_deg\_q239k1 & 53.2 & 1.97 & --- & 1.00 & 1.97 \\
attntest & attn\_dense\_hd128\_local\_noncausal & 103.7 & 1.17 & 1.19 & --- & 0.98 \\
attntest & attn\_varlen\_hd128\_local & 55.5 & 1.60 & 1.20 & 1.01 & 1.33 \\
attntest & attn\_dense\_hd128\_softcap\_gqa & 357.8 & 2.09 & --- & --- & 2.09 \\
attntest & attn\_dense\_hd128\_sink & 112.2 & 1.31 & 1.40 & 0.94 & 0.94 \\
attntest & attn\_varlen\_hd128\_sink & 54.5 & 1.38 & 1.18 & 0.82 & 1.17 \\
attntest & attn\_kv\_hd128\_leftpad & --- & --- & --- & --- & --- \\
attntest & attn\_dense\_hd128\_seq8k & 493.0 & 0.93 & 1.10 & 0.84 & 0.84 \\
attntest & attn\_dense\_hd128\_seq16k & 896.1 & 0.88 & 1.06 & 0.78 & 0.83 \\
edge & attn\_edge\_causal\_q1k1 & 51.6 & 2.19 & 1.84 & 1.09 & 1.19 \\
edge & attn\_edge\_causal\_q3k3 & 51.5 & 2.18 & 1.81 & 1.00 & 1.20 \\
edge & attn\_edge\_causal\_q64k1 & 52.0 & 2.20 & --- & 1.05 & 2.09 \\
edge & attn\_edge\_causal\_q64k32 & 52.0 & 2.18 & --- & 1.05 & 2.07 \\
edge & attn\_edge\_causal\_q64k128 & 51.6 & 2.17 & 1.84 & 1.05 & 1.18 \\
edge & attn\_edge\_causal\_q128k192 & 51.9 & 1.88 & 1.68 & 1.01 & 1.11 \\
edge & attn\_edge\_causal\_q255k256 & 52.6 & 1.89 & 1.65 & 0.99 & 1.14 \\
edge & attn\_edge\_causal\_q108k256 & 52.1 & 2.01 & 1.59 & 1.02 & 1.26 \\
edge & attn\_edge\_causal\_q113k128 & 52.4 & 2.18 & 1.86 & 1.05 & 1.17 \\
edge & attn\_edge\_causal\_q113k211 & 51.7 & 1.90 & 1.60 & 1.02 & 1.18 \\
edge & attn\_edge\_causal\_q128k217 & 51.9 & 1.90 & 1.60 & 1.01 & 1.18 \\
edge & attn\_edge\_causal\_q799k3 & 52.9 & 2.02 & --- & 1.07 & 1.89 \\
edge & attn\_edge\_causal\_q384k256 & 52.5 & 2.04 & --- & 1.02 & 1.99 \\
edge & attn\_edge\_causal\_q640k128 & 52.2 & 2.05 & --- & 1.05 & 1.94 \\
edge & attn\_edge\_causal\_q512k256 & 54.4 & 1.96 & --- & 1.02 & 1.91 \\
edge & attn\_edge\_causal\_q1023k1024 & --- & --- & --- & --- & --- \\
edge & attn\_edge\_causal\_q1024k1023 & --- & --- & --- & --- & --- \\
edge & attn\_edge\_causal\_q2048k2048 & 78.0 & 1.42 & 1.53 & 1.01 & 0.93 \\
edge & attn\_edge\_noncausal\_q1k1 & 52.6 & 2.21 & 1.85 & 1.07 & 1.19 \\
edge & attn\_edge\_noncausal\_q64k128 & 52.6 & 2.26 & 1.83 & 1.07 & 1.23 \\
edge & attn\_edge\_noncausal\_q255k256 & 52.7 & 1.92 & 1.82 & 0.99 & 1.05 \\
edge & attn\_edge\_noncausal\_q128k217 & 52.4 & 2.03 & 1.65 & 1.02 & 1.23 \\
edge & attn\_edge\_noncausal\_q512k256 & 53.2 & 2.04 & --- & 1.01 & 2.01 \\
edge & attn\_edge\_noncausal\_q1024k1023 & 105.5 & 2.74 & --- & 1.67 & 1.63 \\
edge & attn\_edge\_local\_q128k217 & 55.0 & 1.97 & 1.61 & 1.06 & 1.22 \\
edge & attn\_edge\_local\_q255k256 & 55.3 & 1.87 & 1.56 & 1.03 & 1.19 \\
edge & attn\_edge\_local\_q512k256 & 54.1 & 1.79 & --- & 1.01 & 1.76 \\
edge & attn\_edge\_local\_q1024k1023 & --- & --- & --- & --- & --- \\
edge & attn\_edge\_gqa\_q113k211 & 53.5 & 1.94 & 1.64 & 1.03 & 1.18 \\
edge & attn\_edge\_gqa\_q384k256 & 53.0 & 2.06 & --- & 1.03 & 2.00 \\
edge & attn\_edge\_gqa\_q640k128 & 52.5 & 2.04 & --- & 1.05 & 1.94 \\
claim & dense\_c\_d64\_s1k & 227.0 & 1.25 & 1.36 & 1.16 & 0.91 \\
claim & dense\_c\_d64\_s4k & 448.0 & 0.99 & 1.13 & 0.94 & 0.87 \\
claim & dense\_c\_d64\_s8k & 759.4 & 0.92 & 1.12 & 0.89 & 0.83 \\
claim & dense\_c\_d128\_s1k & 153.7 & 1.25 & 1.39 & 1.07 & 0.90 \\
claim & dense\_c\_d128\_s4k & 297.9 & 0.99 & 1.17 & 0.89 & 0.85 \\
claim & dense\_c\_d128\_s8k & 496.8 & 0.92 & 1.09 & 0.82 & 0.85 \\
claim & dense\_c\_d128\_gqa8\_s2k & 184.5 & 1.11 & 1.22 & 0.95 & 0.91 \\
claim & dense\_c\_d128\_gqa8\_s8k & 477.3 & 0.90 & 1.11 & 0.80 & 0.82 \\
claim & dense\_c\_d96\_s2k & 183.4 & 0.85 & 1.25 & --- & 0.68 \\
claim & dense\_c\_d128\_softcap50\_s2k & 802.5 & 1.92 & --- & --- & 1.92 \\
claim & dense\_c\_d128\_sink\_s2k & 194.3 & 1.07 & 1.23 & 0.89 & 0.87 \\
claim & dense\_c\_d256\_s2k & --- & \multicolumn{4}{l}{UNSUP({ours})} \\
claim & dense\_c\_d256\_s8k & --- & \multicolumn{4}{l}{UNSUP({ours})} \\
claim & dense\_c\_d192\_s2k & --- & \multicolumn{4}{l}{UNSUP({ours, FA4, FlashInfer})} \\
claim & dense\_c\_d192\_dv128\_s2k & 161.8 & 1.32 & 1.51 & --- & 0.87 \\
claim & dense\_nc\_d128\_s2k & 271.7 & 0.93 & 1.17 & 0.94 & 0.80 \\
claim & dense\_nc\_d128\_s8k & 889.3 & 0.88 & 1.10 & 0.93 & 0.79 \\
claim & dense\_nc\_d256\_s8k & --- & \multicolumn{4}{l}{UNSUP({ours})} \\
claim & dense\_c\_d64\_fp8\_s8k & --- & \multicolumn{4}{l}{UNSUP({ours, cuDNN, FlashInfer})} \\
claim & dense\_c\_d128\_fp8\_s4k & --- & \multicolumn{4}{l}{UNSUP({ours, cuDNN, FlashInfer})} \\
claim & dense\_c\_d128\_fp8\_s8k & --- & \multicolumn{4}{l}{UNSUP({ours, cuDNN, FlashInfer})} \\
claim & dense\_c\_d128\_fp8\_gqa\_s4k & --- & \multicolumn{4}{l}{UNSUP({ours, cuDNN, FlashInfer})} \\
claim & dense\_c\_d256\_fp8\_s8k & --- & \multicolumn{4}{l}{UNSUP({ours, FA4, cuDNN, FlashInfer})} \\
claim & dense\_c\_d128\_fp8\_scalar\_s4k & --- & \multicolumn{4}{l}{UNSUP({ours, cuDNN})} \\
claim & dense\_c\_d128\_fp8\_scalar\_gqa\_s4k & --- & \multicolumn{4}{l}{UNSUP({ours, cuDNN})} \\
claim & dense\_c\_d64\_fp8\_scalar\_s8k & --- & \multicolumn{4}{l}{UNSUP({ours, cuDNN})} \\
claim & dense\_c\_d128\_swa256\_s4k & 128.9 & 1.05 & 1.14 & 1.01 & 0.92 \\
claim & dense\_c\_d128\_swa4k\_s4k & 266.2 & 0.97 & 0.92 & 0.85 & 0.97 \\
claim & varlen\_c\_d128\_mix & 54.1 & 1.59 & 1.19 & 0.98 & 1.33 \\
claim & varlen\_c\_d64\_mix & 51.9 & 1.77 & 1.18 & 0.98 & 1.50 \\
claim & varlen\_c\_d128\_mqa\_mix & 54.1 & 1.62 & 1.20 & 0.99 & 1.35 \\
claim & varlen\_c\_d128\_realmix & 214.9 & 1.02 & 0.87 & 0.88 & 1.02 \\
claim & llama8b\_dec\_b16\_s512 & 55.1 & 1.56 & 1.81 & 1.12 & 0.86 \\
claim & llama8b\_dec\_b32\_s1k & 77.2 & 1.04 & 1.45 & 1.07 & 0.71 \\
claim & llama8b\_dec\_b16\_s4k & 98.0 & 1.20 & 1.33 & 1.06 & 0.90 \\
claim & llama8b\_dec\_b64\_s4k & 230.9 & 0.95 & 1.18 & 1.12 & 0.80 \\
claim & llama8b\_dec\_b16\_s4k\_p16 & 100.2 & 1.09 & 1.36 & 1.09 & 0.80 \\
claim & llama8b\_dec\_b16\_s4k\_p128 & 96.5 & 1.19 & 1.32 & 1.04 & 0.90 \\
claim & qwen3\_dec\_b16\_s4k & --- & --- & --- & --- & --- \\
claim & qwen3\_dec\_b64\_s4k & 145.1 & 1.02 & 1.18 & 1.12 & 0.87 \\
claim & llama70b\_dec\_b16\_s4k & 98.9 & 1.18 & 1.35 & 1.07 & 0.88 \\
claim & mistral\_dec\_b16\_s4k & 100.0 & 1.22 & 1.35 & 1.07 & 0.91 \\
claim & gemma2\_dec\_b16\_s4k & --- & \multicolumn{4}{l}{UNSUP({ours, FA4, cuDNN, FlashInfer})} \\
claim & llama8b\_dec\_b16\_s16k & 220.8 & 1.05 & 1.20 & 1.09 & 0.87 \\
claim & llama8b\_dec\_b64\_s16k & 844.5 & 0.97 & 1.32 & 1.30 & 0.74 \\
claim & llama8b\_dec\_b16\_s32k & 401.3 & 0.96 & 1.22 & 1.16 & 0.79 \\
claim & qwen3\_dec\_b16\_s32k & --- & --- & --- & --- & --- \\
claim & llama70b\_dec\_b4\_s128k & --- & --- & --- & --- & --- \\
claim & mistral\_dec\_b16\_s16k & 101.4 & 1.18 & 1.36 & 1.08 & 0.86 \\
claim & llama8b\_pf\_q512 & 59.5 & 1.58 & 1.52 & 1.05 & 1.04 \\
claim & llama8b\_pf\_q2k & 85.3 & 1.09 & 1.38 & 0.87 & 0.79 \\
claim & llama8b\_pf\_q8k & 505.9 & 0.97 & 1.17 & 0.88 & 0.83 \\
claim & llama8b\_pf\_q2k\_p16 & 86.4 & 1.07 & 1.36 & 0.87 & 0.79 \\
claim & qwen3\_pf\_q4k & 166.2 & 1.06 & 1.23 & 0.87 & 0.86 \\
claim & llama70b\_pf\_q8k & 989.7 & 0.96 & 1.06 & 0.89 & 0.90 \\
claim & mistral\_pf\_q4k & 169.3 & 1.07 & 0.93 & 0.88 & 1.07 \\
claim & gemma2\_pf\_q4k & --- & \multicolumn{4}{l}{UNSUP({ours, FA4, cuDNN, FlashInfer})} \\
claim & llama8b\_pf\_ragged & 215.5 & 1.02 & 0.86 & 0.88 & 1.02 \\
claim & llama8b\_ext\_q512\_k16k & --- & --- & --- & --- & --- \\
claim & llama8b\_ext\_q1k\_k64k & 1010.2 & 0.97 & 1.10 & 1.04 & 0.88 \\
claim & qwen3\_ext\_q512\_k32k & --- & --- & --- & --- & --- \\
claim & llama70b\_ext\_q1k\_k128k & 4019.9 & 0.94 & 1.07 & 1.07 & 0.88 \\
claim & llama8b\_spec\_q2\_b16\_s1k & 62.9 & 1.19 & 0.35 & 1.07 & 1.11 \\
claim & llama8b\_spec\_q4\_b32\_s4k & 145.1 & 0.93 & 0.14 & 1.09 & 0.85 \\
claim & llama8b\_spec\_q8\_b16\_s4k & 99.8 & 1.00 & 0.17 & 1.03 & 0.98 \\
claim & qwen3\_spec\_q4\_b16\_s4k & --- & --- & --- & --- & --- \\
claim & mixed\_2pf2k\_8dec1k & 141.4 & 0.94 & 0.93 & 0.90 & 0.94 \\
claim & mixed\_1pf8k\_64dec4k & 789.6 & 0.78 & 0.56 & 0.49 & 0.78 \\
claim & mixed\_2pf2k\_8dec16k & 351.1 & 1.17 & 0.52 & 0.65 & 1.17 \\
claim & llama8b\_dec\_bf16q\_fp8kv\_b16\_s1k & 224.4 & 4.69 & 6.20 & --- & 0.76 \\
claim & llama8b\_dec\_bf16q\_fp8kv\_b32\_s4k & 1171.8 & 8.57 & 9.77 & --- & 0.88 \\
claim & llama8b\_dec\_bf16q\_fp8kv\_b16\_s16k & 2088.7 & 9.91 & 11.35 & --- & 0.87 \\
claim & llama8b\_pf\_bf16q\_fp8kv\_q2k & 132.7 & 1.70 & 2.12 & --- & 0.80 \\
claim & llama8b\_dec\_bf16q\_fp8kv\_scalar\_b16\_s1k & 228.8 & 4.55 & 2.96 & 2.42 & 1.54 \\
claim & llama8b\_dec\_bf16q\_fp8kv\_scalar\_b32\_s4k & 1161.5 & 8.30 & 7.28 & 7.23 & 1.14 \\
claim & llama8b\_dec\_fp8\_scalar\_b16\_s4k & --- & \multicolumn{4}{l}{UNSUP({ours, cuDNN})} \\
claim & llama8b\_pf\_fp8\_scalar\_q2k & --- & \multicolumn{4}{l}{UNSUP({ours, cuDNN})} \\
real & attn\_real\_llama8b\_dec\_s512 & 56.6 & 1.50 & 1.36 & --- & 1.10 \\
real & attn\_real\_llama8b\_dec\_s4k & 100.7 & 1.17 & 0.64 & --- & 1.17 \\
real & attn\_real\_llama8b\_dec\_s16k & 224.3 & 1.01 & 0.43 & --- & 1.01 \\
real & attn\_real\_llama8b\_dec\_b64\_s4k & 234.2 & 0.88 & 0.50 & --- & 0.88 \\
real & attn\_real\_qwen3moe\_dec\_s512 & 54.5 & 1.49 & 1.38 & --- & 1.08 \\
real & attn\_real\_qwen3moe\_dec\_s4k & --- & --- & --- & --- & --- \\
real & attn\_real\_qwen3moe\_dec\_s32k & --- & --- & --- & --- & --- \\
real & attn\_real\_llama8b\_pf\_s2k & 128.2 & 1.04 & 0.75 & --- & 1.04 \\
real & attn\_real\_llama8b\_pf\_s16k & 1865.5 & 0.91 & 0.48 & --- & 0.91 \\
real & attn\_real\_qwen3moe\_pf\_s4k & 167.8 & 0.98 & 0.55 & --- & 0.98 \\
\end{longtable}

\normalsize\twocolumn

\section{Trace Statistics and Methodology}
\label{app:traces}

\begin{figure}[t]
\centering
\includegraphics[width=\columnwidth]{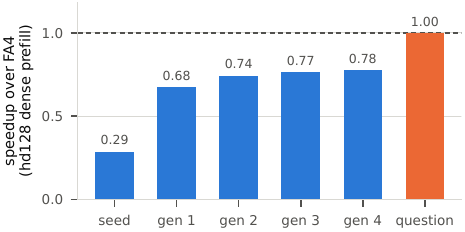}
\caption{Evolutionary search on the Blackwell dense-prefill kernel, then one question. Four
generations of an AVO-style loop (six mutations each, monotone lineage, a strategist writing
hypothesis-shaped directives) took a correct port from 0.29$\times$ to 0.78$\times$ of FA4. The
remaining gap was outside the kernel: a question about where the time was going located a
synchronous host copy of about 38~$\mu$s per call, and removing it brought the dense envelope to
parity.}
\label{fig:evo}
\end{figure}

\begin{figure}[t]
\centering
\includegraphics[width=\columnwidth]{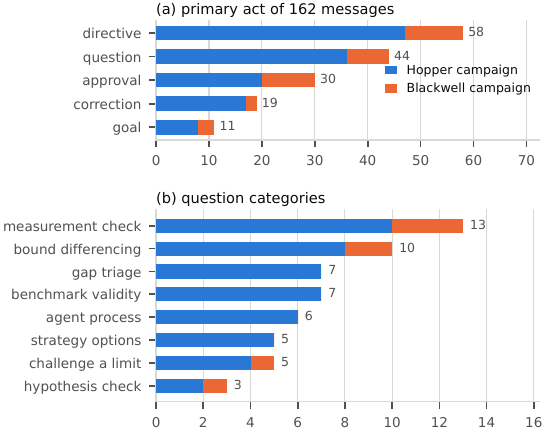}
\caption{The 162 human messages across the 16 \sys sessions. (a) Primary act. (b) Categories of
the 56 messages whose primary act was a question, excluding 8 housekeeping questions about the
repository.}
\label{fig:messages}
\end{figure}

\paragraph{Sources.} Sixteen \sys sessions are archived: nine Claude Code sessions as JSONL
transcripts with their subagent transcripts, and seven Cursor sessions as exported JSONL. Claude
Code transcripts record the model per assistant message (\texttt{claude-opus-4-8} in seven sessions,
\texttt{claude-opus-5} in two); Cursor exports do not. The equipped agent's run is a checkpoint
directory with one record per candidate (source, build, correctness, timing, promotion decision,
timestamp) and the orchestrator's log, memory file, and knowledge store.

\paragraph{Parsing.} A tool call is an assistant content block of type tool use, de-duplicated by
its identifier across the parent transcript and any compaction summaries; a subagent's calls are
counted separately and attributed to the parent session. Human turns are user messages that are not
tool results, Stop-hook injections, or system reminders; in Cursor exports they are identified by
the user record's timestamp tag. Tool calls are classified by regular expressions over the tool
name and arguments into: profiler runs, SASS or PTX inspections, builds, benchmark or evaluator runs,
documentation reads, sanitizer runs, and reads of the agent's own kernel files; the categories
overlap where a shell command does two things. The ``tool-result text re-reading own kernel'' figure
of Table~\ref{tab:h2h} is the share of tool-result characters returned by file reads whose path is
one of the kernel sources, over all tool-result characters in the Claude Code parent transcripts
(26\%; for the equipped agent, the same quantity over its per-step tool logs, 82\%).

\paragraph{Labeling.} One author labeled each of the 162 human messages by its primary act (goal,
directive, question, approval, correction), flagged whether it was phrased as a question, and
assigned the 44 question-act messages and the 12 question-shaped corrections to the categories of
\S\ref{sec:workflow-questions}; eight further questions concerned repository housekeeping and are
excluded from Figure~\ref{fig:messages}(b). The agent-side counts (35 structured questions to the
human, 24 offering to accept or stop, 126 limit assertions, 59 Stop-hook returns) are from the Claude
Code sessions only, where these events are structured; the Cursor exports do not separate them.
Figure~\ref{fig:messages} shows the resulting distribution. The labels, the parsing scripts, and the
per-session inventory are released with the traces.

\section{Baseline Agent Details}
\label{app:baseline}

\begin{table}[t]
\centering\small
\caption{What \baseline supplied to the model at every step.}
\label{tab:baseline}
\setlength{\tabcolsep}{4pt}
\begin{tabular}{@{}p{0.34\columnwidth}p{0.6\columnwidth}@{}}
\toprule
Component & Distilled output in the prompt \\
\midrule
Lever selector & one of 16 optimizations, keyed to the NCU bottleneck label \\
NCU parser & bottleneck diagnosis for the worst configuration \\
SASS analyzer & instruction-mix deltas vs.\ the champion and a CUTLASS oracle \\
Doc retrieval & BM25 over the PTX ISA, CUDA guide, NCU and Nsight docs, CUTLASS docs, Hopper tuning guide \\
Reference libraries & hints extracted from DeepGEMM, ThunderKittens, CUTLASS (1.6~GB on disk) \\
Knowledge store & typed notes (kernel, profiling, algorithmic), 15 seeded \\
Memory & causal lessons from recent steps: change, hypothesis, outcome, cause \\
Sanitizer, anti-cheat & bounded compute-sanitizer runs; forbidden-symbol checks \\
\bottomrule
\end{tabular}
\end{table}

Table~\ref{tab:baseline} lists the components of \baseline and what each contributed to the step
prompt. The per-step prompt assembled, in order: the task statement and hard rules; the chosen lever
with its rationale; the NCU diagnosis for the worst configuration; the SASS instruction-mix deltas
against the champion; the retrieved documentation sections; the reference-library hints; the
knowledge-store notes matching the lever; the memory file's recent lessons; the current champion
source; and the instruction to write one improved candidate to a fixed path and stop. The median
prompt was 22.6K characters excluding the champion source, 37\% of it retrieved or pre-fetched
material. Promotion required a correct candidate whose worst-configuration ratio exceeded the
champion's by more than a 3\% noise band. The plateau detector fired after eight steps without
promotion and opened a rewrite branch seeded at the lineage's pre-plateau knee; 37 branches were
opened, 36 from the same step. Three human rollbacks discarded 117 steps. The run's 313 kept steps,
the 133 model-driven ones among them, the memory file, and the knowledge store are released with
the specification from which the agent was built.

\section{Benchmark Methodology}
\label{app:bench}

\paragraph{Timing.} Each scenario is timed on one GPU with every implementation run back to back
from byte-identical inputs. After 20 warm-up calls, the harness takes three trials of 40 CUDA-event
samples each, flushing a 32~MB buffer through L2 and synchronizing before every sample, and reports
the median of the 120 samples; the per-scenario speedup is the upstream median over the
implementation's median. Clocks are not locked, which is why only ratios are reported. cuDNN and
FlashInfer are given the same packed and paged buffers as our kernel and the upstream, and a
scenario is marked unsupported for an implementation when it rejects the configuration or fails
correctness.

\paragraph{Correctness.} Every candidate's outputs are compared with the upstream kernel's and with
an fp32 reference on every scenario, with a per-dtype tolerance; a candidate that fails any scenario
scores zero. In the head-to-head, the equipped agent's gate additionally required token-for-token
equality with FA3 on a greedy decode inside vLLM.

\paragraph{Repeats and noise.} The Hopper suite was run as three full repeats in each of four
independent batches on the 8$\times$H200 node, one scenario per GPU at a time, with a single-GPU
serial control that agrees with the parallel runs to within 0.8\%. Per-scenario medians over the
three repeats are the reported numbers; the four batches agree to within 0.3\% on every aggregate,
and per-scenario ratios vary by 1\% at the median and 6\% at most. The Blackwell suite was run once
on the 8$\times$B200 node with the same protocol. The \texttt{edge} scenarios are short kernels for
which dispatch overhead is a visible share of wall-clock time, so their ratios overstate kernel-only
gains; the \texttt{claim} and \texttt{real} sets are long enough that dispatch is negligible.

\end{document}